\documentclass{article}

\usepackage{arxiv}

\usepackage[utf8]{inputenc} 
\usepackage[T1]{fontenc}    
\usepackage{url}            
\usepackage{booktabs}       
\usepackage{amsfonts}       
\usepackage{nicefrac}       
\usepackage{microtype}      
\usepackage{lipsum}
\usepackage{graphicx}
\graphicspath{ {./images/} }
\usepackage{nomencl}
\usepackage{lineno}
\usepackage{xcolor}
\usepackage{amsmath}
\usepackage{amssymb}
\usepackage[linesnumbered,ruled,vlined]{algorithm2e}
\SetKwInput{KwInput}{Input}
\usepackage{verbatim}
\usepackage{caption}
\usepackage{float}
\usepackage{upgreek}
\usepackage{verbatimbox}
\usepackage{stackengine}
\usepackage{subfig}
\usepackage{subfloat}
\usepackage{moreverb,url}
\usepackage[english]{babel}
\usepackage[T1]{fontenc}
\usepackage[colorlinks,bookmarksopen,bookmarksnumbered,citecolor=red,urlcolor=red]{hyperref}
 
\newcommand\BibTeX{{\rmfamily B\kern-.05em \textsc{i\kern-.025em b}\kern-.08em
T\kern-.1667em\lower.7ex\hbox{E}\kern-.125emX}}

\title{Traffic Management of Autonomous Vehicles using Policy Based Deep Reinforcement Learning and Intelligent Routing}

\author{
 Anum Mushtaq \\
  Department of Computer and Information sciences\\
  Pakistan Institute of Engineering and Applied Sciences\\
  Islamabad, Pakistan\\
   \And
 Muhammad Azeem Sarwar \\
  Department of Computer and Information sciences\\
  Pakistan Institute of Engineering and Applied Sciences\\
  Islamabad, Pakistan\\
  \And
 Irfan ul Haq \\
  Department of Computer and Information sciences\\
  Pakistan Institute of Engineering and Applied Sciences\\
  Islamabad, Pakistan\\
  \And
 Asifullah Khan \\
  Department of Computer and Information sciences\\
  Pakistan Institute of Engineering and Applied Sciences\\
  Islamabad, Pakistan\\
  \And
 Omair Shafiq \\
  School of Information Technology\\
  Carleton University\\
  Ottawa, Canada\\
}

\begin{document}
\maketitle
\begin{abstract}
Deep Reinforcement Learning (DRL) uses diverse, unstructured data and makes RL capable of learning complex policies in high dimensional  environments. Intelligent Transportation System (ITS) based on Autonomous Vehicles (AVs) offers an excellent playground for policy-based DRL. Deep learning architectures solve computational challenges of traditional algorithms while helping in real-world adoption and deployment of AVs. One of the main challenges in AVs implementation is that it can worsen traffic congestion on roads if not reliably and efficiently managed. Considering each vehicle's holistic effect and using efficient and reliable techniques could genuinely help optimise traffic flow management and congestion reduction. For this purpose, we proposed a intelligent traffic control system that deals with complex traffic congestion scenarios at intersections and behind the intersections. We proposed a DRL-based signal control system that dynamically adjusts traffic signals according to the current congestion situation on intersections. To deal with the congestion on roads behind the intersection, we used re-routing technique to load balance the vehicles on road networks. To achieve the actual benefits of the proposed approach, we break down the data silos and use all the data coming from sensors, detectors, vehicles and roads in combination to achieve sustainable results. We used SUMO micro-simulator for our simulations. The significance of our proposed approach is manifested from the results. 
\end{abstract}


\section{Introduction}
Continuous progress in the automotive industry has made fully automated systems a reality that can handle the whole driving process independently without requiring any human interruption. Recent advancements in cutting edge technology promise to help in the prevention of accidents by avoiding unsafe lane changes, drifting into adjacent lanes, generating warnings to vehicles while backing up, and automatic braking systems when a vehicle suddenly stops or slows down \cite{czech2018autonomous}. A combination of advanced software and hardware technologies are used in these vehicles to achieve the desired output. Fully autonomous vehicles advance from level 0 ( fully controlled by driver, no automation) to level 5 (no driver presents, full automation) \cite{vagia2016literature} \cite{hancock2019future}.  The level 5 vehicles require a complex set of algorithms for different functionalities such as path planning, scene recognition \cite{buluswar1998color}, object recognition and tracking etc.\cite{kato2015open}, \cite{fujiyoshi2019deep}.
\par Today there are many severe challenges faced by the transportation industry, one of them is the high density of vehicles in critical areas that leads to traffic congestion which is directly associated with an increase in carbon dioxide emissions, unwanted delays, noise, parking issues \cite{gomez2001parallel}, accidents and unnecessary fuel consumption \cite{zambrano2019centralized}. These situations become further complicated when there are vehicles with no human drivers present inside. To address these concerns, there is a dire need for efficient solutions for traffic management and control. These solutions should be capable of generating benefits such as reducing delays, travel times, congestion management, and environmental pollution \cite{carp2018autonomous}. In the context of AVs, it is imperative to carefully analyze traffic demand and predict future traffic conditions so that more accurate and effective route optimization is done. These methods could be very effective for mitigating the adverse effects of traffic congestion and enabling city management entities by improving traffic flow.
\par There have been many AI techniques which have been used in the last few years for traffic flow and control \cite{namazi2019intelligent}, \cite{mushtaq2021traffic1}, \cite{khayyam2020artificial}. However, Reinforcement Learning (RL) has caught the particular interest of the research community working on autonomous vehicles. The most important quality of RL is beguiling-- a method of dealing agents by reward and punishment schemes without going into details of how results will be achieved \cite{kaelbling1996reinforcement}. DRL is the extension of the RL that is poised to revolutionize the artificial intelligence industry. DRL plays a significant role in building autonomous systems and previously intractable scale problems. The algorithms of DRL provide a higher-level understanding of autonomous systems allowing control policies to be directly learned by the inputs of the real-world \cite{arulkumaran2017deep}. DRL algorithms are beneficial to solve the traffic flow problems such as congestion, delays and unnecessary emissions by increasing the efficiency of existing infrastructure systems. One of the essential systems to improve is the Traffic Signal Controllers (TSC) because one of the most critical areas of congestion is the traffic signals, where a situation could worsen if not properly managed, especially in AVs.
\par In this paper, we focused on the development of intelligent TSC on the intersections by taking advantage of recent AI advancements using DRL techniques for reducing congestion, minimizing delays, and reducing queue lengths. As the road intersections are meant to disperse traffic flow, they can easily lead to obstacles in traffic flow and cause accidents. Statistics reveal that more than one-third of total traffic delays are due to delays on the intersections and thus lead to more than 50 per cent of the total accidents \cite{yu2013study}. Analyzing intersections, exploring efficient methodologies for traffic management on intersections and developing intelligent traffic signal controls have immense significance. Taking these measures can improve traffic conditions by increasing traffic efficiency while ensuring traffic safety.  
To address these concerns, an intelligent approach is proposed in this paper for improving traffic flow on intersections and balancing traffic while minimizing delays. Following are the contributions of this paper:
\begin{itemize}
 \item We are proposing a DRL-based framework for traffic signal management in complex traffic environments. A smart traffic signal controller takes inputs from various sensors and is trained using deep reinforcement learning. The intelligent signal optimally selects the appropriate signal sequence that should be enacted according to the current state of intersection in the dynamically changing complex simulation environment. 
   
    \item We also propose a technique for load balancing the traffic by rerouting the vehicles on the road network to alternate paths. On the one hand, congestion at the intersection is controlled by intelligent signal whereas on the other hand traffic coming behind the intersections is rerouted to other paths between their Origin-Destinations to reduce the congestion at the intersection and behind the intersection, thus smoothing traffic flow.  
\end{itemize} 
We used SUMO (a tool for traffic simulations) \cite{krajzewicz2012recent} for simulating the road networks, traffic generation and the infrastructure.
The organization of next sections is as follows: In Section II we showed the related work in the current area. In Section III, we give a detailed explanation of our proposed approach. In Section IV experimental setup is presented while Section V presents the results and detailed discussions on them. Finally Section VI concludes the paper

\section{Literature Review}
This study constitutes different fields of study including reinforcement learning, autonomous vehicles, routing and traffic flow. Earlier studies on traffic control using RL were limited due to the lack of computational power and simple simulations \cite{Thorpe96trafficlight}, \cite{Wiering00multi-agentreinforcement}, \cite{brockfeld2001optimizing}, \cite{abdulhai2003reinforcement}. Continuous improvements have been made in this area in early 21st century and variety of realistic and complex simulations have been developed. Recently many AI techniques have been used in traffic signal controls such as Deep Q-learning [\cite{li2016traffic}, \cite{mousavi2017traffic}], Q-learning \cite{liao2009study}, \cite{liu2017distributed}, and fuzzy \cite{askerzade2010control}. Traffic researchers use micro-simulators which are the most popular tools for producing real world behaviour of traffic and model distinct entities for individual vehicles. The type of reinforcement learning, reward definition, state/action space definition, traffic simulator, vehicle generation model and traffic network geometry used is different in different studies. In previous studies state space is defined as some traffic's attribute and the most popular attributes are traffic flow \cite{balaji2010urban}, \cite{arel2010reinforcement} and number of queued vehicles \cite{abdulhai2003reinforcement}, \cite{Wiering00multi-agentreinforcement}, \cite{abdoos2013holonic}, \cite{chin2011q}. All available signal phases define the action space \cite{arel2010reinforcement}, \cite{6502719} or action space could be restricted to only green phase \cite{balaji2010urban}, \cite{chin2011q}, \cite{abdoos2013holonic}. Change in queued vehicles and change in delay are most common definitions used for rewards \cite{balaji2010urban}, \cite{chin2011q}, \cite{abdoos2013holonic}. A comprehensive review of traffic signals controlled by reinforcement learning is given in \cite{el2014design} and \cite{mannion2016experimental}.
\par Technological advances in this area ensure efficient transportation systems, and on an unprecedented scale, a large amount of varied data could be collected. This high-quality data could be used with minimal abstraction as in \cite{genders2016using} authors proposed a control system for traffic lights that make use of this type of data using deep reinforcement learning. They proposed a discrete state encoding, which input the deep convolutional neural net and new state space. Q-learning algorithm is used with experience replay. The cumulative delay, average travel time and average queue length decreased effectively. They used traffic simulator SUMO for their simulations. In \cite{van2016coordinated} authors investigate the optimized learning policies for traffic signals. They control traffic light problems by combining the coordinated algorithm with the Q-learning algorithm and giving a new reward function. Their approach reduces the travel time effectively and minimizes the possible causes of instability in reinforcement learning. In \cite{askerzade2010control}, fuzzy logic is used to calculate the traffic light's optimum extension time. Two sensors are used to identify traffic flow and extend the time membership function used by the fuzzy light controller. In the fuzzy model a dynamic environment is analyzed, and model regularly change itself to adapt this environment. Changing the model repeatedly and applying it to adjust with the environment consumes huge amount of processing power and eventually system's performance degrades. 
\par In \cite{chin2011q}, traffic congestion is reduced by minimizing queue length, and fixed signals are used for turning signal green. The duration of green time is extended based on the queue lengths. Its performance proves better than fully fixed signals. In \cite{cabrejas2020assessment} different reward functions are used for comparing the performance of agents. The authors used real-world constraints to simulate a junction such as realistic controllers, sensor inputs, green times, and calibrated demand and stage sequencing. Their reward time was based on queue length, time spent, junction throughput, average speed, lost time etc., and their performance measure depended on average waiting time. Their results showed that, across all demand levels, speed maximization causes the lowest average waiting time. In \cite{lin2018efficient} authors proposed a traffic control strategy using a deep reinforcement learning algorithm. Their algorithm tunes parameters and relaxes the need for fixed traffic demand assumptions.   
\par Managing traffic flow is an essential part of the Intelligent Transportation System (ITS), but many challenges need to be addressed. Because of the variable and complex traffic dynamics, model-based control methods do not perform well. Model-free data-driven techniques such as RL are more suitable methods for such situations. The DRL method is good for signal management, but for the efficient traffic flow and congestion reduction, only traffic signal controllers are insufficient. All the studies mentioned above only focus on traffic signal controls; however, there is a need to use some hybrid techniques to maintain the overall traffic flow.  Using a combination of techniques,  the congestion on the intersection and the whole network could be managed efficiently, and a better traffic management system could be developed. To achieve this goal, in our previous study \cite{mushtaq2021traffic}, we proposed a technique using value-based reinforcement learning and smart rerouting, which showed very effective results. In the current study, we used policy-based DRL methodology to balance the traffic on the road infrastructure.

\begin{figure*}[ht]
    \centering
    \includegraphics[width= 13cm,height=10cm]{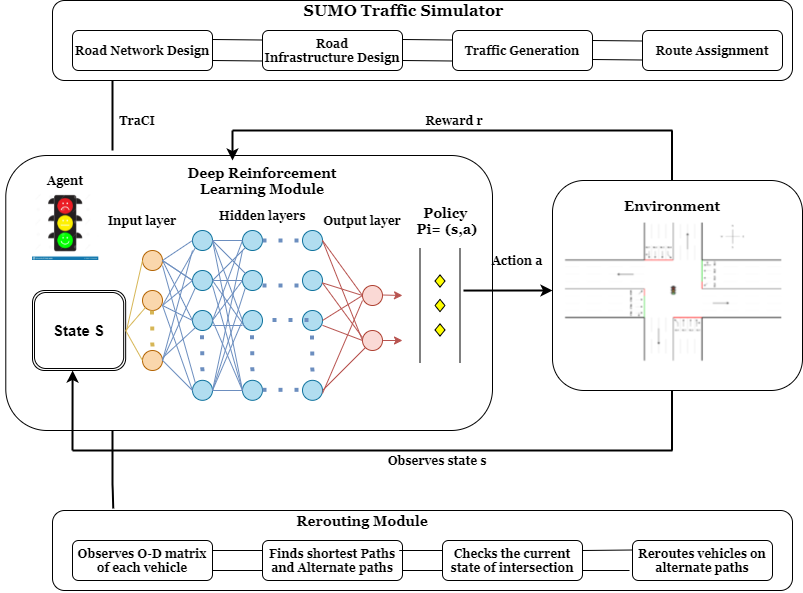}
    \caption{\textsf {Intelligent Traffic Signal Management and rerouting [Proposed Approach]}}
    \label{fig:Proposed Methodology}
\end{figure*}

\section{Proposed Approach}
AVs are a long-standing goal of AI, and navigation of these vehicles requires highly efficient models and infrastructure. Although there have been many developments in this area for the last two decades, there is still a considerable need for intelligent systems to replace experienced human drivers. Successful navigation of autonomous vehicles from one point to another is not sufficient enough for the AVs; instead, we also would require intelligent algorithms, smart infrastructure and proper congestion management techniques in future.
We present an intelligent policy iteration based DRL algorithm for traffic signal control. This intelligent algorithm dynamically control signals based on the current situation at the intersection and resolves congestion. It reduces the long wait times, long queues and delays at the intersections, thus making traffic lights efficient and resolving congestion. We then applied the rerouting technique to the traffic behind the intersection. The vehicles coming behind that have yet not reached the intersection do not have to stuck and delayed due to the congested intersection. By checking the traffic congestion on the intersection, vehicles behind the intersections are rerouted to other routes based on their origin-destination matrix if there are vehicles more than the threshold limit. The graphical representation of our proposed approach is shown in Fig \ref{fig:Proposed Methodology}.   

\subsection{States}
According to our approach, the traffic signal phase, vehicles speed and position represent states. The current state of the vehicles could be obtained by the V2I communication using sensors deployed on the road network .  We divided the road into small segments, and each segment has some sensors embedded on it, which returns true in the presence of the vehicle and false in its absence. 

\subsection{Actions}
Actions are the set of functions performed by the traffic lights. Traffic light becomes green for those vehicles whose waiting time and queue length are greater than others. Actions are performed by looking at the current state of the intersection dynamically.  

\subsection{Reward}
One of the significant and most important part of the RL is reward as it shows the output of the specific action in a specific state. The subsequent actions should be selected very carefully as they are dependent on the output of the reward function. Reward contain normative content, stipulating the goal of the agent and could be negative or positive. In our approach, we used vehicle's waiting time  as a reward in a specific lane.  Waiting time of each newly entered vehicle is recorded and total time of all the vehicles in a specific lane is added to calculate commutative waiting time. The vehicle that leaves the lane, its waiting time is not added to the total delay time. Our global objective is the overall improvement and management of traffic flow and  congestion reduction, and to achieve this, we use a local objective function. The reward is calculated as old waiting time - new waiting time. Therefore, the reward will be negative if the current waiting time is more than the previous and the reward will be positive if the current waiting time is less than the previous waiting time.
\newline In our proposed approach, we modelled our problem as MDP, where agents interact with the environment, learns from it, and based on its learning it take decisions. The objective is to maximize the performance by learning optimal policies that are best suitable at that instance of time. We could represent an MDP as a set \( S,A,T, \gamma,I,R\) here \(S_t\) are the states, \(A_t\) represents the actions, \(T\) represents the transition function that shows the probability of transition between states,\( \gamma \) is discount factor, \(I\) is the initial state's distribution and \(R_t\) represents the reward.
 if the reward and probability are unknown than MDP can be treated as RL's problem. In RL environment, an agent in state \(s_{t_{i}}\,\in \,S_t\), at each time \(t_{i}\), takes an action \(a_{t_{i}}\,\in \, A_t \) depending on observation and then transition function \(T\) leads it to new state \(s_{t_{i}+1}\,\in\, S_t \).  The reward \(r_{t_{1}}\,\in\,R_t \) could  be maximized by learning an optimal policy \(\pi: S_t\times A_t \rightarrow [ 0, 1] \). The probability of selecting an action in any state represents its optimal policy. The expected cumulative discounted future reward  can be calculated as
 \begin{equation}
    R_{t}=E\Bigg[\sum_{k=0}^{\infty} \gamma^k \: r_{t+k}\Bigg]
    \end{equation}
Here \( \gamma \) is used as trade-off between exploration vs exploitation.

If \(x\) denotes the result on the output layer than we can say that
\[
    f(x)= 
\begin{cases}
    x,& \text{if } x\geq 1\\
    0, & \text{otherwise}
\end{cases}
\]

In RL learning, the optimal policy is beneficial for continuous and high dimensional spaces, which require very high computational power and memory. Defining policies contain a manageable set of parameters that are less consuming and efficient.
In our case, the problem is to optimize traffic signals intelligently to determine what action to take at state (s) to maximize reward. This objective could be obtained by fine-tuning its vectors of parameters \(\theta_i\) and selecting the best action for policy \(\pi\).

\begin{equation}
    \pi(a_i|s_i,\theta_i) = P_{r}\big\{ A_{t} = a_i | S_{t} = s_i, \; \theta_{t} = \theta_i\}
    \end{equation}
    
It can be defined as the probability of an action \(a_i\) at state \(s_i\) is the policy \(\pi\) with \(\theta_i\) parameter.

\begin{figure*}[ht]
    \centering
    \includegraphics[width= 12cm,height=8cm]{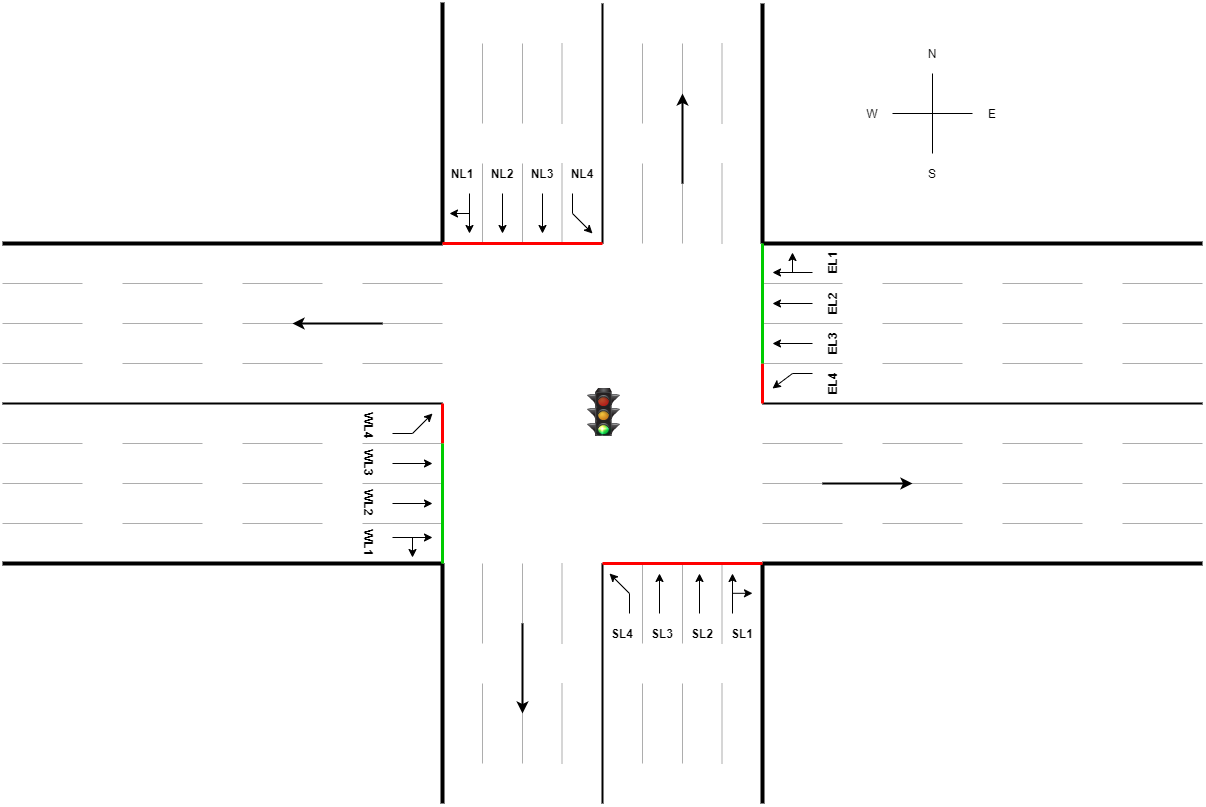}
    \caption{\textsf {Traffic light Intersection used for simulation}}
    \label{fig:intersection}
\end{figure*}

\subsection{System training and objective function}
There could be several objective functions used while managing traffic optimization problems like balancing queue lengths, increasing throughput, minimizing the number of vehicles at intersections etc. Minimizing the delay is the main objective of our agent, that is directly proportional to maximizing
traffic throughput and minimizing queue length in this research. To maximize reward based on \(\theta_i\), we will define our objective function as:

\begin{equation}
    J(\theta_i) \dot{=} V_{\pi_{\theta_i}} (s_0)
    \end{equation}
    
Where \(V_{\pi_{\theta_i}}\) is value function, \(\pi_{\theta_i}\) is policy and \(s_0\) is initial state. It indicates that maximizing \(J(\theta_i)\) means maximizing \(V_{\pi_{\theta_i(s_i)}}\) 

\begin{equation}
    \nabla J(\theta_i) = \nabla V_{\pi_{\theta_i}} (s_0)
    \end{equation}
    
  According to the policy gradient theorem \cite{sutton1999policy}

\begin{equation}
    \nabla J(\theta_i) \propto \sum_{s_i} d(s_{i})\sum_{a_i} q_{\pi} (s_{i},a_{i}) \nabla\pi(a_{i} |s_{i}, \theta_i)
    \end{equation}
    
Here \(d(s_{i})\) is the distribution under \(\pi\) which means  the probability of being at state \(s_{i}\) while following policy \(\pi\). Where \(q(s_{i}, a_{i})\) is the action value function under \(\pi\), and \(\nabla\pi(a_{i} |s_{i}, \theta_i)\) is gradient of \(\pi\) given state and \(\theta_i\). 

\par
\begin{algorithm}[H]
\caption{Training of Neural Network based on policy gradient method}
\DontPrintSemicolon
  
  \KwInput{\emph{Initialize parameters i.e. \(\pi(s_i|\theta_{i}^{\pi})\) randomly  }  \newline
    \emph{Initialize step counter 0, \(t\gets 0\)  } \newline
    \emph{Initialize Memory's Reply Buffer RB  }}
  \KwData{\emph{Initialize set of sample trajectory policy \(S_{0}, A_{0}, R_{0}, S_{1}, A_{1}, R_{1},..., S_{T}\)}}
  
  \While{Epochs$=$ 1 $;$ Epochs $<$ Total Epochs $;$ Epochs$++$}
    {
    \emph{Start the Simulation counter with 1st Step t} \newline
    \emph{Initialize observation state \(s_{0}\)}\newline
    \emph{ Set \(t_{start} = t\)} \newline
        \For{\(t \gets 0\; to\; t\)}
            {
            \emph Perform action $a_{i} = \pi(s_i|\theta_{i}a^{\pi})$\;
            \emph Analyze reward $r_{t}$ and next state $s_{i+1}$\;
            \emph{Update transition values $(s_{0}, a_{0}, r, s_{n})$ in RB}
            }
    \emph{ Sample small batches from RB}\;
    \emph{Define Objective function: $J(\theta_i) \dot{=} V_{\pi_{\theta_i}} (s_0)$}\;
    \emph{Calculate probability for actions at specific state by: $\nabla J(\theta_i) \propto \sum_{s} d(s_{i})\sum_{a} q_{\pi} (s_{i},a_{i}) \nabla\pi(a_{i} |s_{i}, \theta_i)$}\;
    \emph{Define expected sum of rewards as: $J(\theta_i) = E\Bigg[\sum_{t=0}^T R_t\,(s_{i}\,,a_{i})\: ; \pi_{\theta_i}\Bigg] = \sum \uptau \; P(\uptau\, ;\, \theta_i) \: R_t(\uptau)$}\;
    \emph{Maximize reward according to:\; $\underset{\theta_i} max J(\theta_i) = \underset{\theta_i} max \sum\uptau \; P(\uptau\, ;\,\theta_i)\: R_t(\uptau)$}\;
    \emph{Do optimization as: $\nabla_{\theta_i} J(\theta_i) = \frac{1}{m}\sum_{i=1}^{m} \sum_{t=0}^{T}\nabla_{\theta_i}\log\pi_{\theta_i}(a_{i}|s_{i})(Q(s_{i} , a_{i})-V_{\varnothing}(s_{i})$}
}
\end{algorithm}

\begin{figure*}[ht]
    \centering
    \includegraphics[width= 12cm,height=10cm]{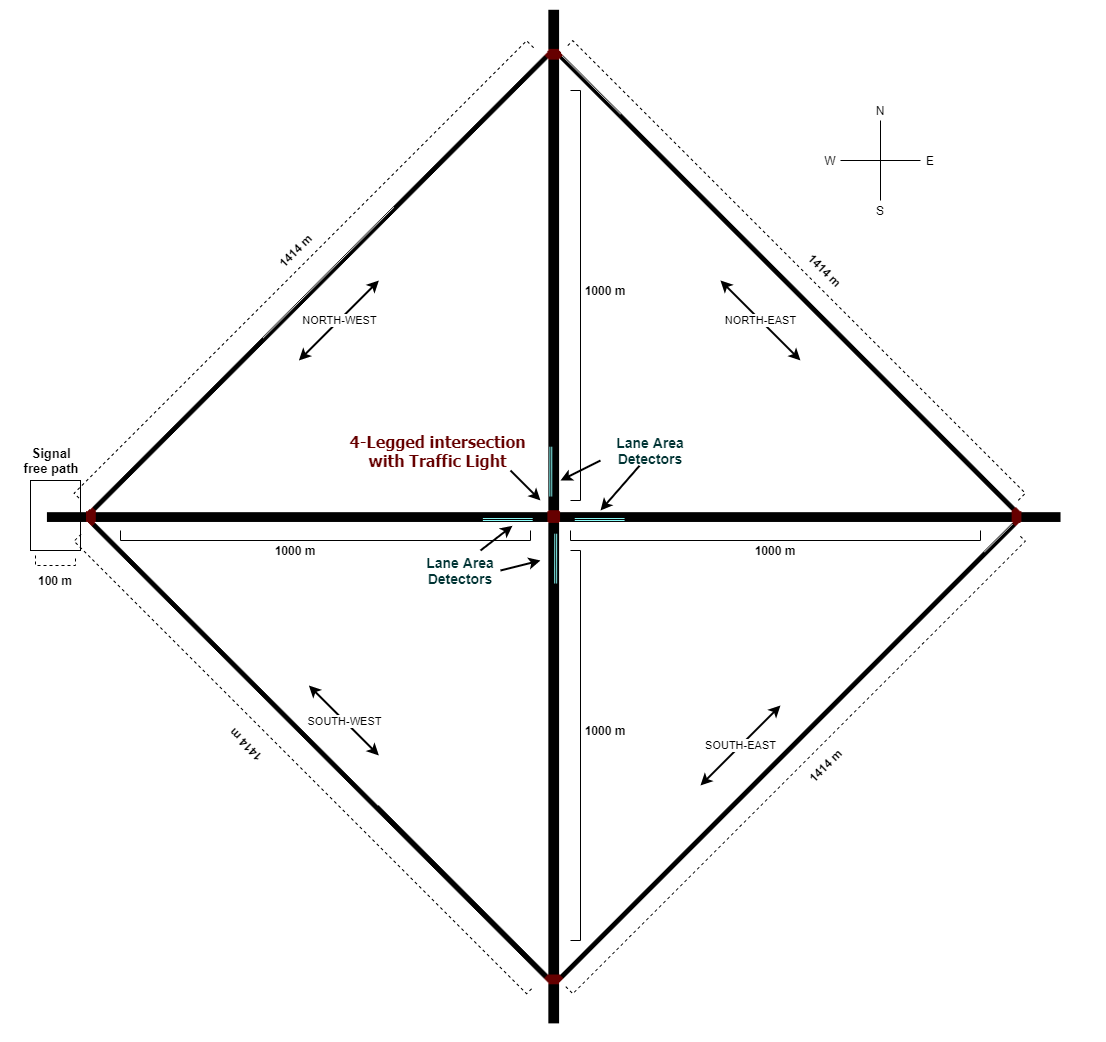}
    \caption{\textsf {Directions of road network}}
    \label{fig:road network}
\end{figure*}

As the objective is to maximize the reward, so for every episode \(T\), starting from 1 to the maximum number of episodes, we define \(J(\theta_i)\) as the expected sum of rewards \(R_t(s_i, a_i)\), following the policy \(\pi(\theta_i)\). 
So if we take episode as trajectory \(\uptau\) from 1 to \(T\), then the expected reward is the summation of all the trajectories of the probability as per \(\theta_i\), \(\uptau\) times the return of this trajectory \(R_t(\uptau)\).

\begin{equation}
    J(\theta_i) = E\Bigg[\sum_{t=0}^T R_t\,(s_{i}\,,a_{i})\: ; \pi_{\theta_i}\Bigg] = \sum \uptau \; P(\uptau\, ;\, \theta_i) \: R_t(\uptau)
    \end{equation}
    
Here, our aim is to find such parameter set \(\theta_i\) that maximizes our reward \(J(\theta_i)\).

 \begin{equation}
  \underset{\theta_i}max J(\theta_i) = \underset{\theta_i}max \sum\uptau \; P(\uptau\, ;\,\theta_i)\:R_t(\uptau)
    \end{equation}
    
Taking derivative and doing mathematical calculations from policy based gradient, we got the following objective function for our traffic light optimization problem:
 
 \begin{equation}
 \triangledown_{\theta_i} J(\theta_i) = \frac{1}{m}\sum_{i=1}^{m} \: \sum_{t=0}^{T}\:\triangledown_{\theta_i}\,\log\pi_{\theta_i}(a_{i}\,|\,s_{i})\:R_t(\uptau^i)
    \end{equation}

In the above equation, \(J(\theta_i)\) is the objective function, say minimizing the cumulative wait time of vehicles, \(m\) is the total episodes in the simulation, \(\pi\) is the policy dependent on the \(\theta_i\) such that if we change the  \(\theta_i\) it will affect the policy as well. When vehicles are at the intersection waiting for the light phase to turn green, then the policy indicates the probability of which light phase will be activated in the current state. \(\uptau^i\) is the ith episode, \(R_t(\uptau^i\)) is the return (total reward) of  \(\uptau^i\), and \(T\) is the number of step in  \(\uptau^i\).  
We can summarize the above equation as the average of all \(m\) trajectories is \(J(\theta_i)\), and each trajectory is the sum of episodes. At each episode, we take the derivative of the log of the policy and multiply it with reward \(R(\uptau^i\)).  
Here, an issue that occurs with \(R(\uptau^i\)) is that when our traffic light agent knows nothing about the environment at the start of the simulation, it takes random actions that may cause an increment in the negative reward. We want to minimize negative rewards during the learning process, and after many episodes, when our reward becomes 0, the \(J(\theta_i)\) will be 0. In this situation, our neural network model will not learn anything new. To avoid this problem, we use discounted rewards.

 \begin{figure*}[ht]
    \centering
    \includegraphics[width= 12cm,height=8cm]{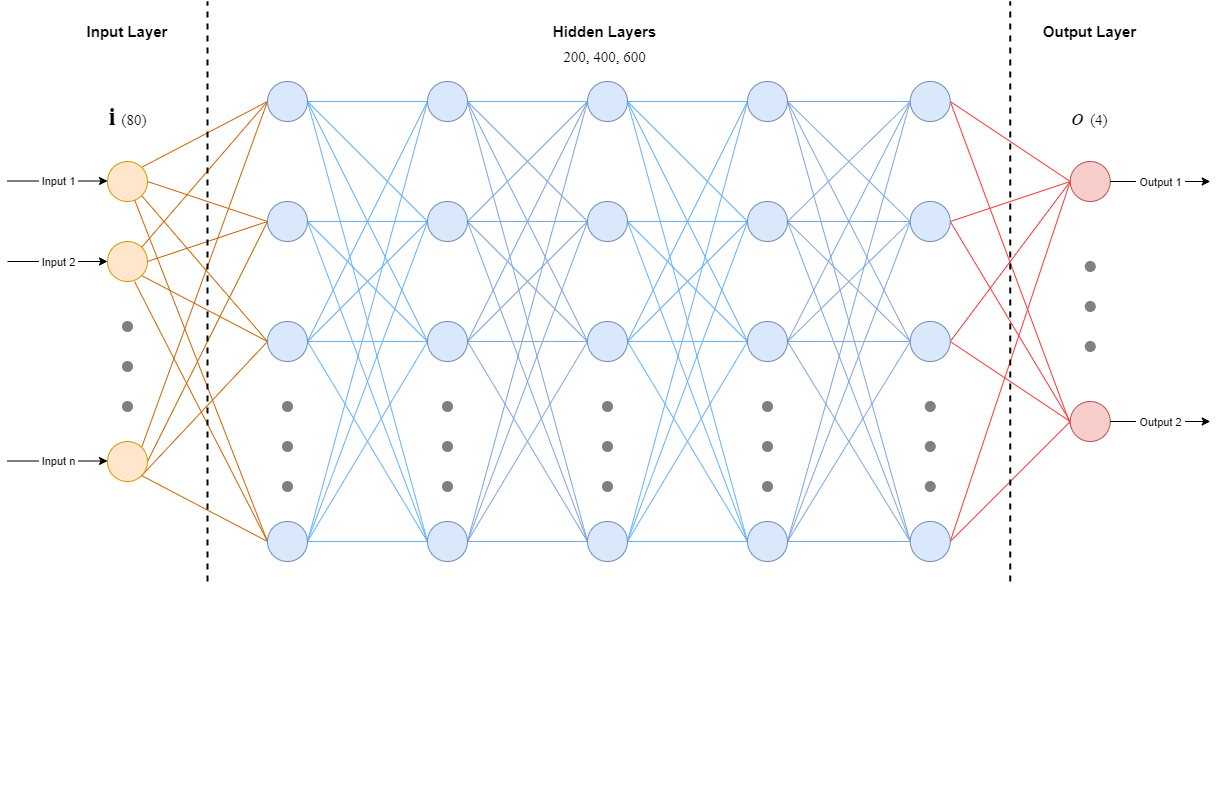}
    \caption{\textsf {Architecture of Neural Network Model Used}}
    \label{fig:nn architecture}
\end{figure*}

\begin{equation}
R_{t} = \sum_{t'}^{T} \:= t\,\gamma^{t'}\:rt'
    \end{equation}
    
Now the equation 8 becomes

\begin{equation}
 \triangledown_{\theta_i} J(\theta_i) = \frac{1}{m}\sum_{i=1}^{m} \: \sum_{t=0}^{T}\:\triangledown_{\theta_i}\,\log\pi_{\theta_i}(a_{i}\,|\,s_{i})\:R_t
    \end{equation}

The \(R_{t}\) in the above equation will return the rewards when we take certain action at step \(t\), but to know which reward value is optimal or which value is not good, we need some baseline or reference point with which we can compare our results. For this purpose, we will take an average of the results of all actions and use it as a reference point. we will use \(Q(s_{i},a_{i})\) that is the value of the specific \(a_i\) at \(s_i\), and \(V(s_{i})\) is the average of all rewards at \(s_i\).   
Using these values, we can rewrite the above equation as
   
    \begin{equation}
    \begin{split}
 \triangledown_{\theta_i} J(\theta_i) = \frac{1}{m}\sum_{i=1}^{m} \: \sum_{t=0}^{T}\:\triangledown_{\theta_i}\,\log\pi_{\theta_i}(a_{i}\,|\,s_{i})\:(Q(s_{i}\, , a_{i}) \\ -\:V_{\varnothing}(s_{i}) )
    \end{split}
    \end{equation}
    
 In the above equation, \(\pi(a_{i}|s_{i})\) tells us which light phase to activate and \(Q(s_{i},a_{i})\:-\:V(s_{i})\) tells us how good our action was.

\subsection{Rerouting of vehicles}

\par
\begin{algorithm*}
\caption{Vehicle's Rerouting }
\emph{Initialize the Memory parameter M}\\
\emph{Set vehicles with their IDs such as}\\
  \emph{ $V_{i}= [{v_{1},v_{2},v_{3},....,v_{n}}] $ }\\
  \emph{Calculate O-D against each vehicle ID }\; 
  \emph{Find all Routes $[R_{i}]$ against each vehicle O-D i.e }\\
  \emph{$R_{i}= [{r_{1},r_{2},r_{3},....,r_{n}}] $ }\\
  \emph{Calculate shortest Routes using Dijkstra algorithm}\\
  \emph{Sort the routes in ascending order}\\
  \emph{Get the vehicle's density at intersection }\\
  \emph {Compute each route's Travel time $(T_{time})$ }\\
    \emph{Store $T_{time}$ in route file}\\
    \ForEach{$V_i$ at Intersection I}{
      \emph{Observe the current state of I}\\
     \emph{Calculate Total-Wait-Time $(T_{wt})$ at I }\\
      {
      \emph{Add the $T_{wt}$ to the $T_{time}$ of shortest path }\\
      \emph{Calculate Updated-Total-Wait-Time $(U_{twt})$}\\
      }
        \If{$U_{twt} > T_{time}$ of alternate routes} 
        {
        Reroute Vehicles to the alternate routes
        }
        \uElseIf{($U_{twt} \leq$ alternate routes)}
        {Stay on the same route}
    }
\end{algorithm*}
\begin{figure*}[htbp]
  \centering
    \subfloat[ 8 Layered Neural Network Model\label{fig:deep}]{\includegraphics[height=16em,width=.45\linewidth]{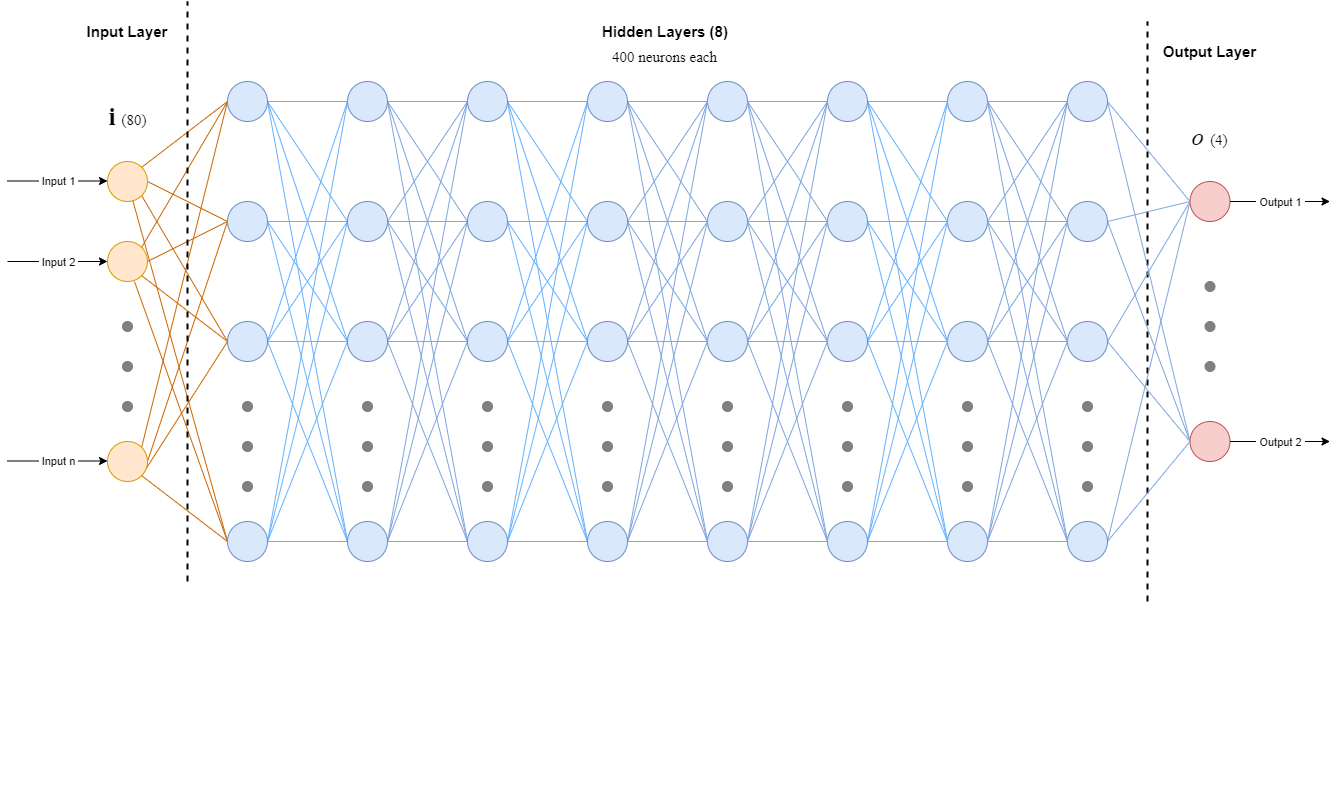}}
    \hfil
    \subfloat[3 Layered Neural Network Model\label{fig:shallow}]{\includegraphics[height=16em,width=.45\linewidth]{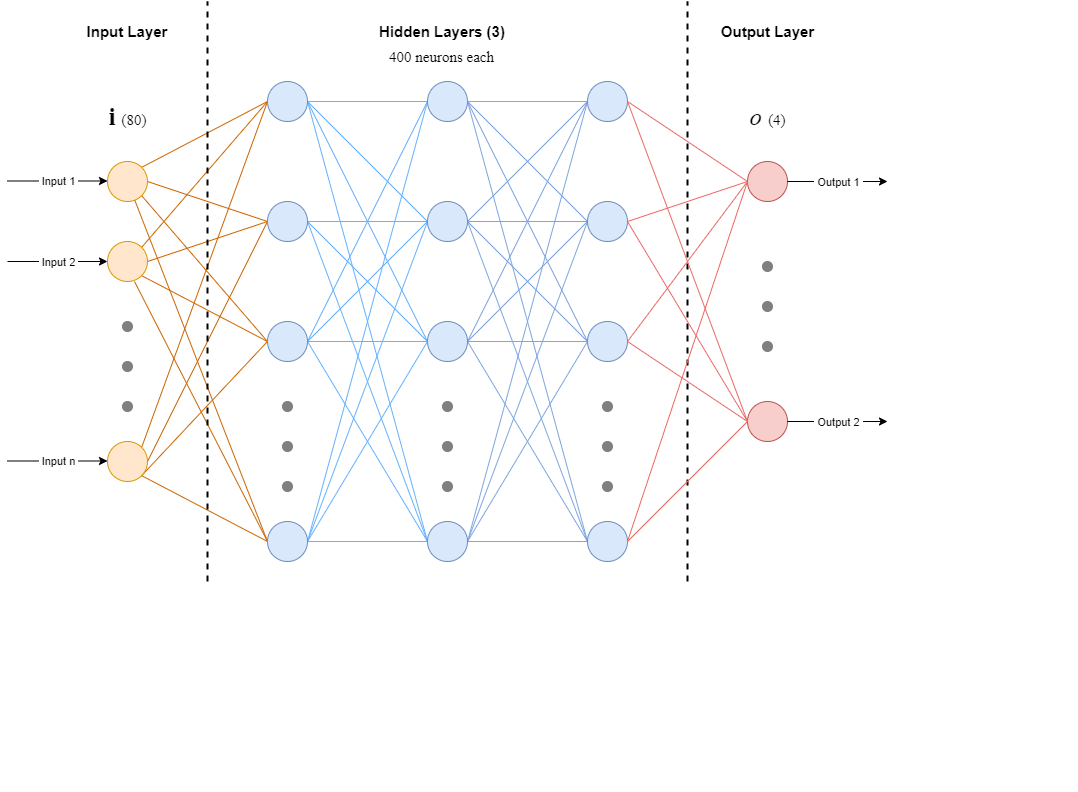}}
    \caption{Comparisons of Neural Networks: Deep vs Shallow}
    \label{fig:deep vs shallow}
\end{figure*}

Traffic on the intersection is managed by RL based intelligent traffic lights that reduce the waiting times of vehicles at the intersection thus reducing long queue lengths. The main purpose of this paper is to optimize the flow of traffic while minimizing the delay. To optimize traffic flow, Vehicles that are stuck in the congestion in long queues behind the intersection are rerouted to less congested paths to load-balance the traffic on the intersection during peak hours. These alternate paths are computed from the perspective of each vehicle's Origin-Destination (OD).
Lane area detectors are used to detect the traffic information on a specific road segment. The traffic data from these detectors is maintained in a separate log file and analyzed every thirty(s) and a critical density limit is set. When the density of vehicles exceeds that threshold limit, the vehicles on the road segment behind the intersection are rerouted to the alternate path leading to the vehicle's destination. The detectors give information about the direction, number of vehicles and their speed on the road segment. In this way, traffic is managed well at the intersection and behind the intersection, providing a solid metaphor for resolve congestion and optimize routes.

\section{Simulation Setup}
The detail of simulation experiments is discussed in this section.
\subsection{Road Network Used}
The length of the road network is shown in Fig.\ref{fig:road network}, that is 1100 meters that we used in our simulations. In the road network, multiple paths are used to reach from one point to other in order to simulate the rerouting of the vehicles in case of congestion. In Fig.\ref{fig:intersection}, we can see the intersection used for traffic signal control. We used one route containing traffic intersection with traffic lights while all other routes are signal-free. The length of the road network from left, right, north and south towards the intersection are 1100 (1000+ 100) meters, while the road on diagonals is 1414 meters on each side. These diagonals are used as an alternate path which is mostly signal free.  
The roads connected to the intersection contain four lanes (4 incoming, 4 outgoings); two middle lanes for vehicles that move towards the left and right lane for vehicles that go towards left and right.

\subsection{Route Assignment in Network}
We used the following two methods for assigning routes to the vehicles.
\begin{enumerate}
    \item Static route assignment: Static assignment is done before starting the simulation. These are initial fixed routes in which 60\% of vehicles will use straight paths through the intersection, and the remaining 40\% will use signal-free routes based on directions that are normally lengthier than signalled routes.    
    \item Dynamic route assignment: Routes are assigned dynamically using TraCI in SUMO, analysing the current state of the traffic. If vehicles' density crosses the threshold limit on certain areas, then dynamically, vehicles are rerouted towards alternate routes.   
\end{enumerate}

\subsection{Detectors used}
 We used four detectors on each lane's side to get the current information of vehicles on the road. The total number of detectors used are 16 for our experiments, and after every 30 seconds, the data of detectors are analyzed. Detectors are placed on south, east, west, and north, and the data of these detectors is stored separately in a log file of each side. When vehicles on the road exceed certain threshold limit, these detectors send a warning message to vehicles coming from the roads, behind the intersection, to wait or change their route according to the rerouting strategy. The direction and speed, along with vehicle's number determined by these detectors. We used this information to shift traffic to other routes where traffic is low in volume.      
 
 \begin{table}[H]
 \centering
 \addstackgap[8pt]{\begin{tabular}{ |p{3cm}||p{3cm}|}
 \hline
 \multicolumn{2}{|c|}{Simulation Parameters} \\
 \hline
 Parameter & Value\\
 \hline
Transmission Range &  1100 (m) \\ \hline
Simulation Time &  9000 (s) \\ \hline
Vehicles per Episode & 4000 \\ \hline
Simulation Time &  39 (min) \\ \hline
Simulation Map &  4 leg intersection \\\hline
Total traffic lights  & 01 \\\hline
Simulator & SUMO\\
 \hline
\end{tabular}}
\end{table}
 
 \begin{figure*}[ht]
  \centering
    \subfloat[Cumulative Delay at the intersection\label{fig:delay1}]{\includegraphics[height=14em,width=.45\linewidth]{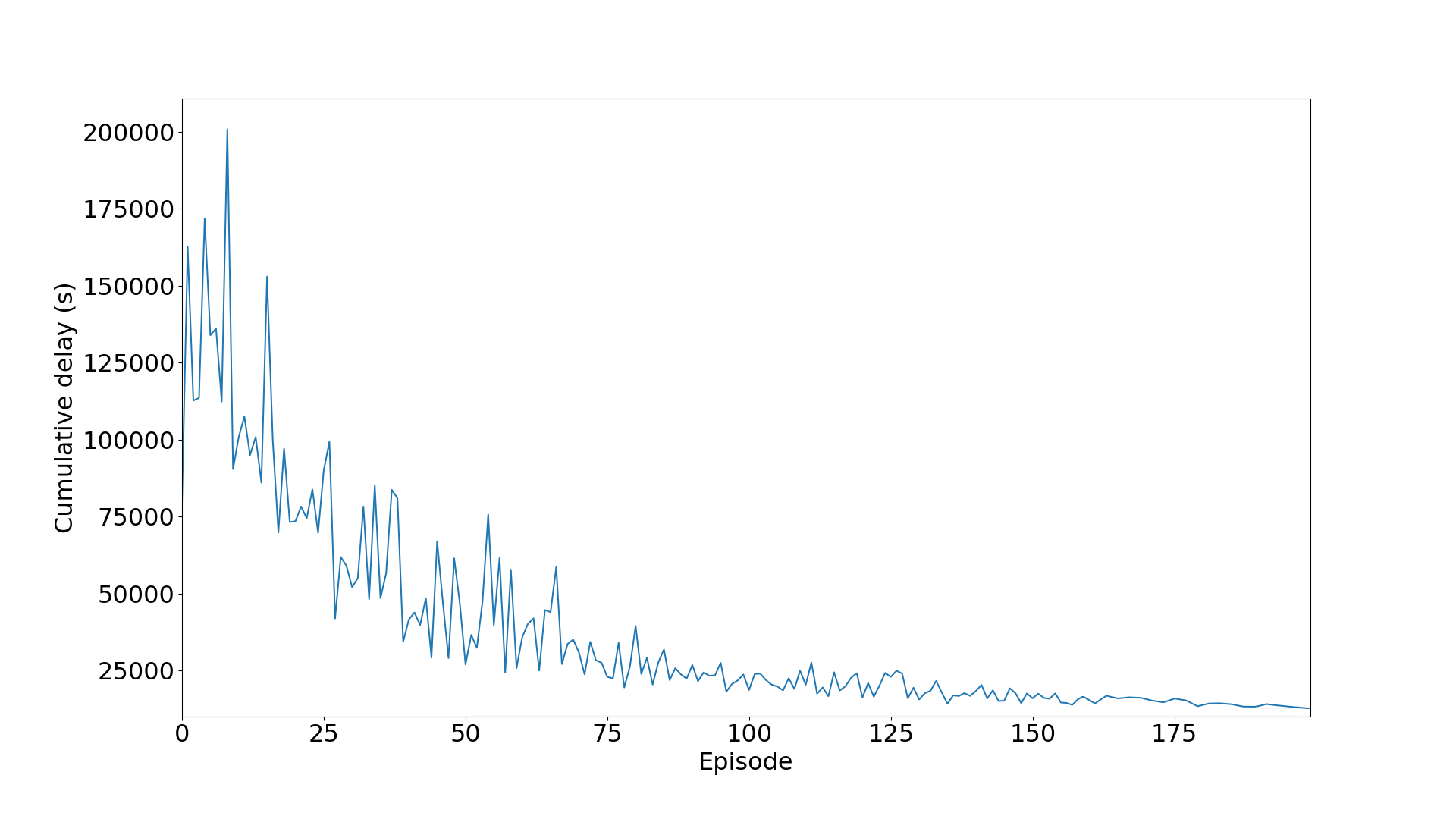}}
    \hfil
    \subfloat[Queue Lengths (average) at the intersection\label{fig:queue1}]{\includegraphics[height=14em,width=.45\linewidth]{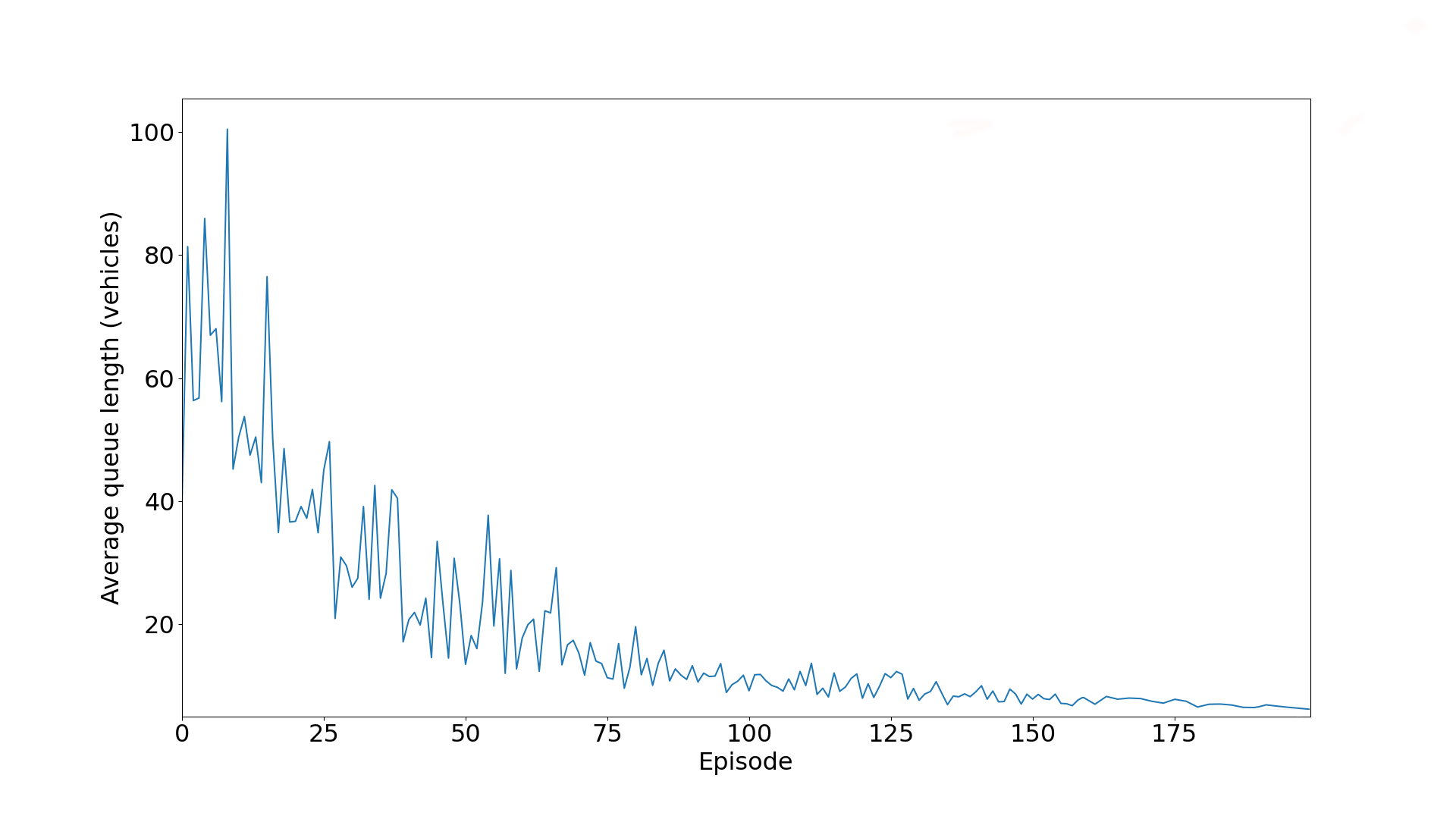}}

    \subfloat[Cumulative Negative Reward \label{fig:reward1}]{\includegraphics[height=14em,width=.45\linewidth]{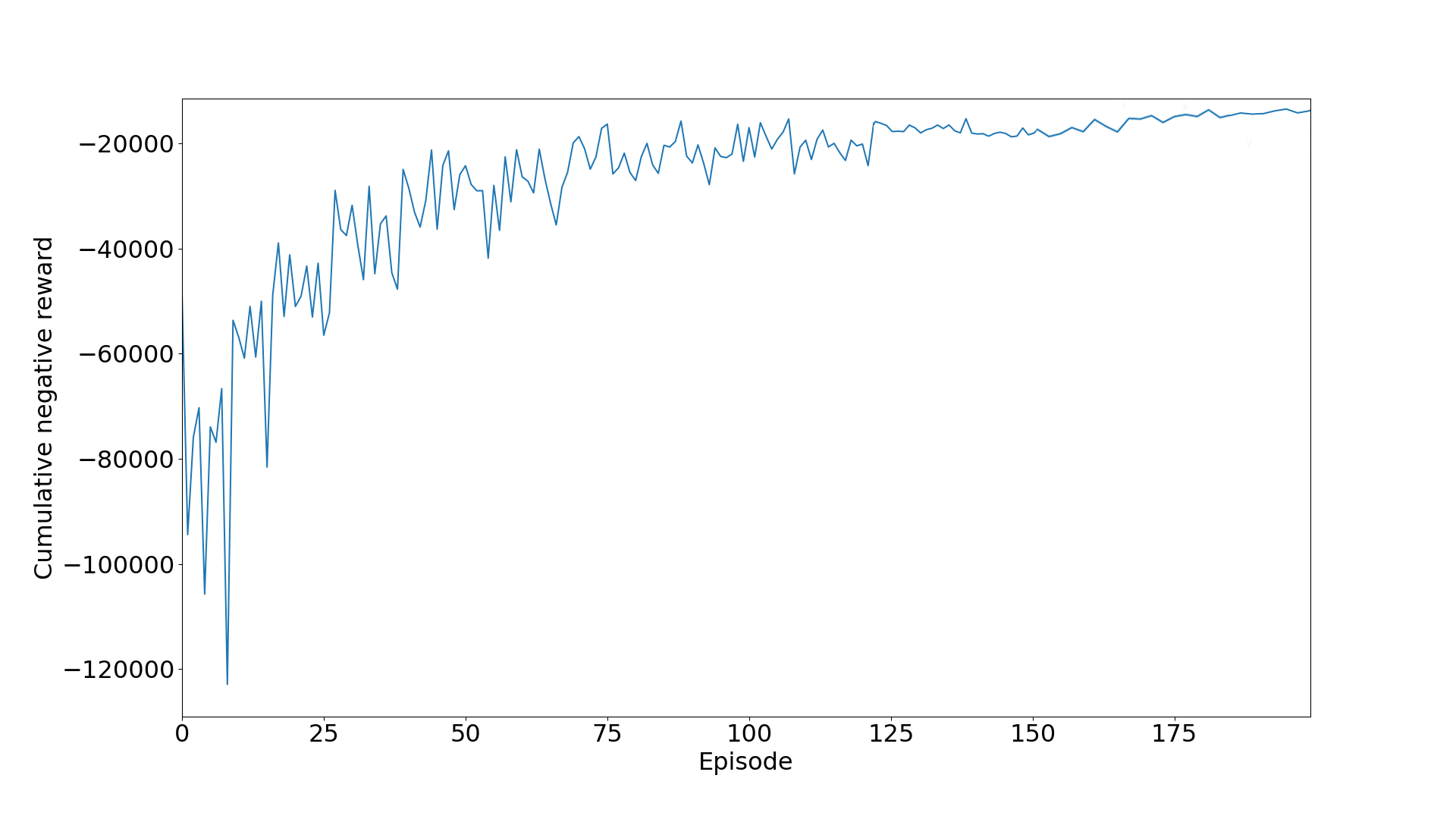}}
    \caption{Results of average delay, queue lengths and rewards for NN1}
    \label{fig:nn model1}
\end{figure*}  

 \subsection{Traffic generation}
 We used almost 4000 different types of vehicles for our simulations, such as cars, trailers, buses and ambulances having varying speeds. Each vehicle has its O-D matrix, based on which route is assigned to vehicles. At the start, all vehicles are assigned the shortest route from their origin to destination. However, if the shortest route is congested, these vehicles could be rerouted to alternate paths from their origin to destination, which could be longer but faster.      
 Our simulation used our intersection route as the shortest path for 60\% of the vehicles and every vehicle has to take the route containing intersection  while other 40\%  vehicles depend on the intersection's situation. If the intersection is congested, then these vehicles are shifted towards an alternate route. In our proposed approach, 60\% traffic is controlled using intelligent traffic signals based on reinforcement learning. In contrast, the remaining 40\% of vehicles are managed by rerouting to alternate paths. These alternate paths may be longer than the shortest route but could be the fastest routes depending on the situation.

\subsection{Sensors used in network}
Different types of sensors are used for getting the current traffic situation on the roads. We used 80 sensors embedded on different points on lanes for incoming traffic. Each side of the four-legged intersection has 20 sensors, among which 10 sensors are used on three lanes, i.e. one right lane and two straight lanes, because these lanes use the same signal. The left lane uses the remaining 10 sensors. The sensors are deployed at appropriate distance from each other, however the sensors' density is higher near the traffic signal. Each sensor has a unique id and provides information about the position of vehicles according to the lanes. A sensor transmits information when vehicles come in its vicinity. The information from these sensors is used to track the vehicles that are heading towards the intersection.    
\begin{table}[H]
\centering
\addstackgap[8pt]{\begin{tabular}{ |p{3cm}||p{3cm}|}
 \hline
 \multicolumn{2}{|c|}{Training Parameters} \\
 \hline
 Parameter & Value\\
 \hline
Total Episodes & 200 \\\hline
Batch Size &  200 \\\hline
Memory Size &  4500 \\ \hline
Number of Actions &  4 \\\hline
GUI   &  True\\\hline
Green phase time & 4(s) \\\hline
Yellow phase time & 2(s) \\\hline
Gamma Value    & 0.50 \\\hline
Number of states &  80 \\ \hline
Maximum Step & 2500\\
 \hline
\end{tabular}
}
\end{table}

\begin{figure*}[ht]
  \centering
    \subfloat[Cumulative Delay at the intersection \label{fig:delay2}]{\includegraphics[height=14em,width=.45\linewidth]{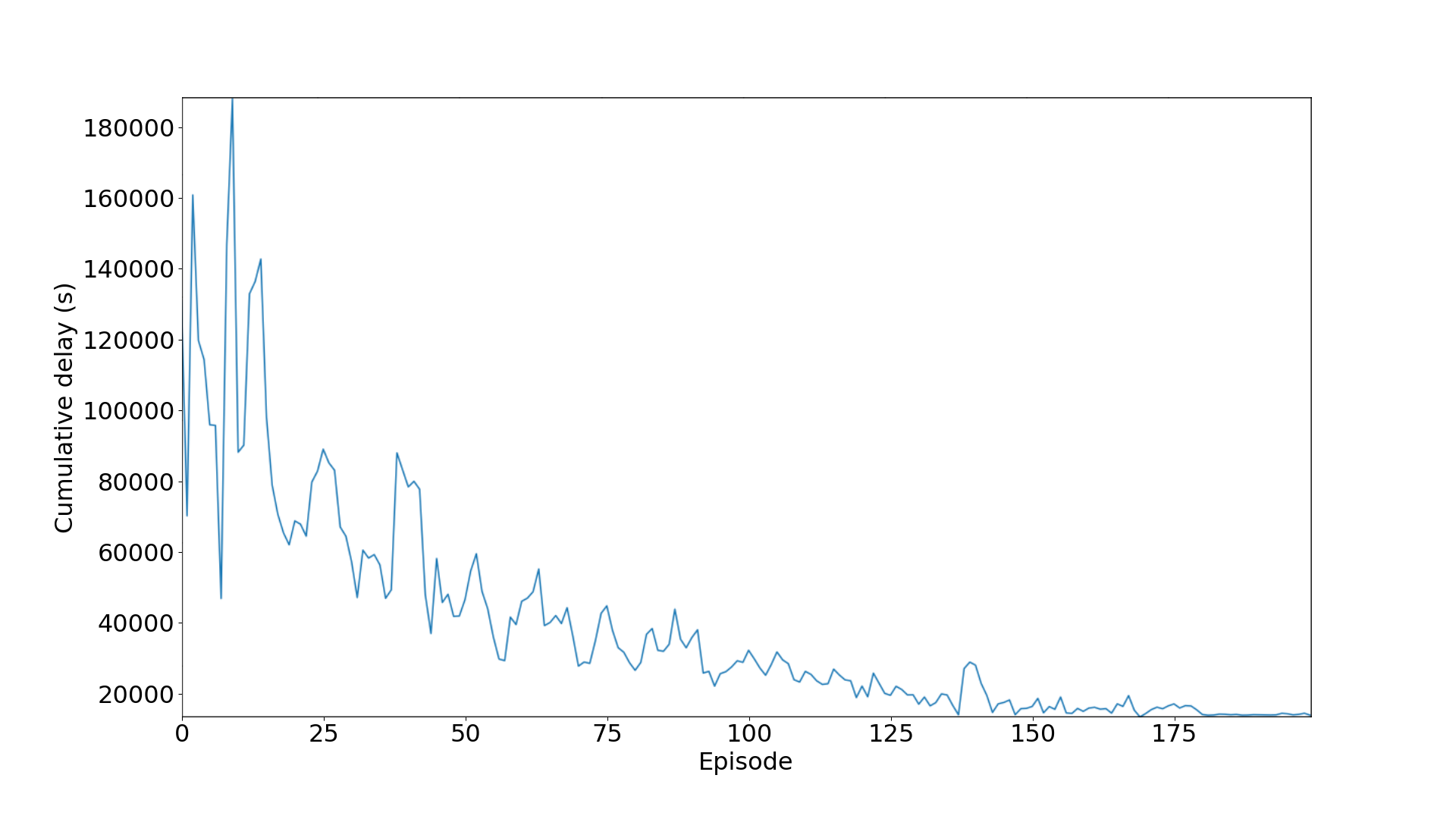}}
    \hfil
    \subfloat[Queue Lengths (average) at the intersection\label{fig:queue2}]{\includegraphics[height=14em,width=.45\linewidth]{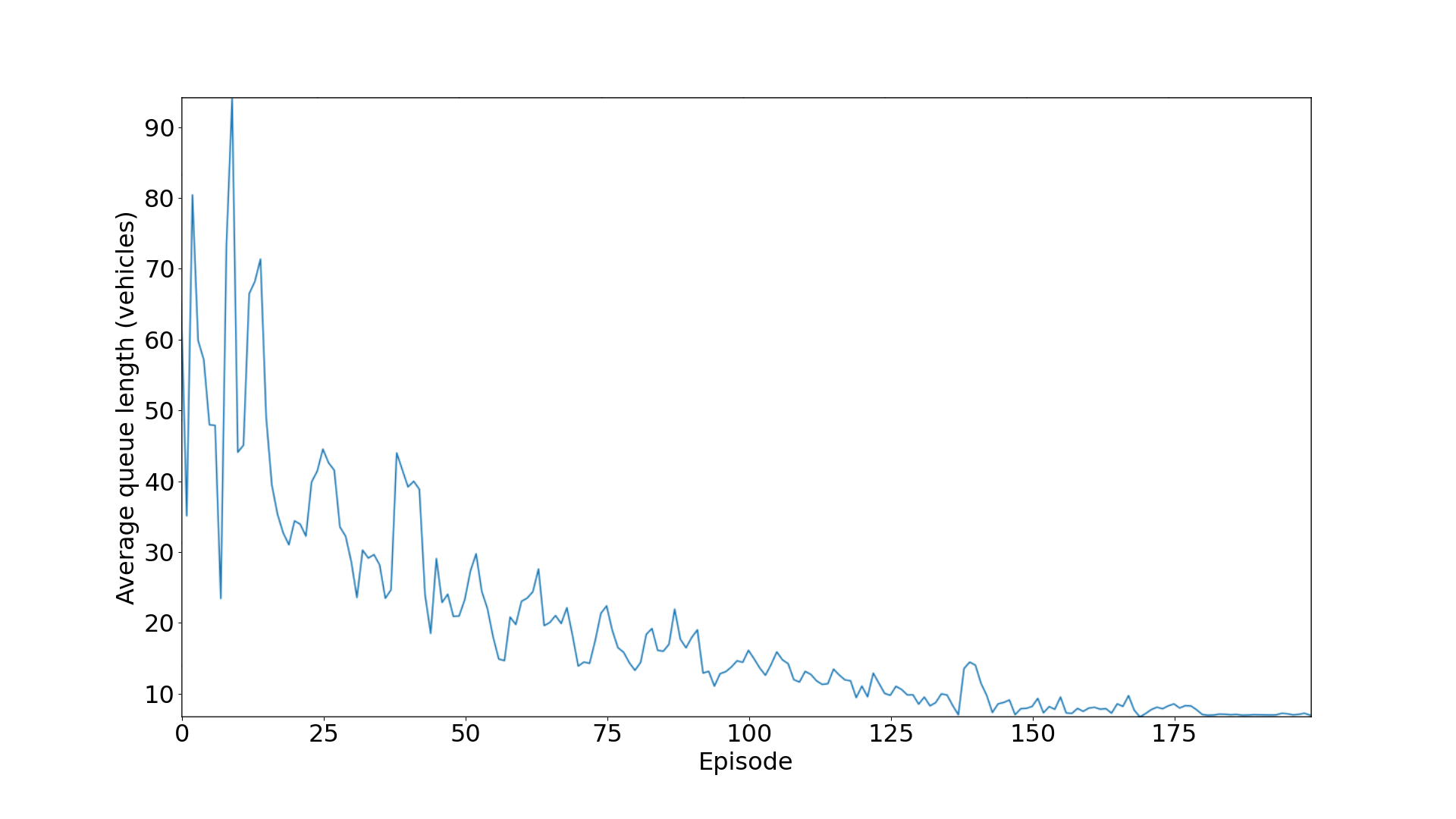}}

    \subfloat[Cumulative Negative Reward \label{fig:reward2}]{\includegraphics[height=14em,width=.45\linewidth]{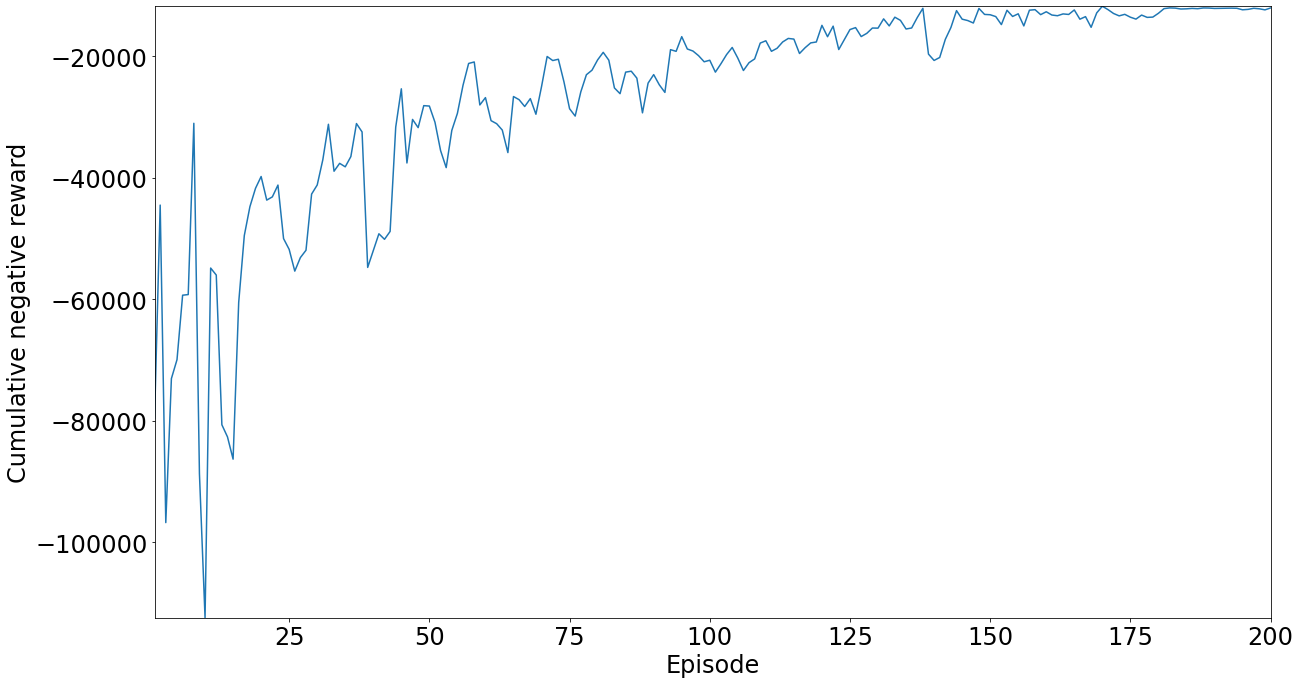}}
    \caption{Results of average delay, queue lengths and rewards for NN2}
    \label{fig:nn model2}
\end{figure*} 

\subsection{Performance Parameters}
The parameters used for performance evaluation are as follows:
\subsubsection{Reward}
The first and the most important parameter used for the performance evaluation is the reward function. Two types of rewards could be uses, i.e. positive and negative. The agent aims to maximize the positive reward or minimize the negative reward. In the presented study, we are minimizing the negative reward and evaluating results based on that. The negative reward tending towards zero means that our agent is learning gradually.   

\subsubsection{Delay}
During peak hours, when the intersection is crowded with heavy traffic and vehicles get stuck in jams, delay increases. After applying our approach, the cumulative delay starts to decrease, which shows improved performance. 

\subsubsection{Queue Length}
The queue length is another critical parameter that shows congestion during high traffic peaks. The long vehicle's queues shows the critical situation at the intersection. The performance parameter used is named as "average queue length" and performance is measured as, when queue lengths start decreasing, it means performance is increasing.

\subsubsection{Simulation Time}
Another critical parameter is the simulation time for all episodes. The simulation stops as the last car leaves the road segment. The maximum time for simulation is 9000(s), and this time will decrease with the improvement in the performance.
\begin{table}[H]
\centering
\addstackgap[8pt]{\begin{tabular}{ |p{3cm}||p{3cm}|}
 \hline
 \multicolumn{2}{|c|}{Neural Network's Structure} \\
 \hline
 Parameter & Value\\
 \hline
NN layers & 200, 400, 600 \\ \hline
 Activation Function &  Relu \\ \hline
 Loss Function    & Mean Squared Error \\ \hline
Optimization & Sigmoid Activation\\
 \hline
\end{tabular}}
\end{table}

\begin{figure*}[ht]
  \centering
    \subfloat[Cumulative Delay at the intersection\label{fig:delay3}]{\includegraphics[height=14em,width=.45\linewidth]{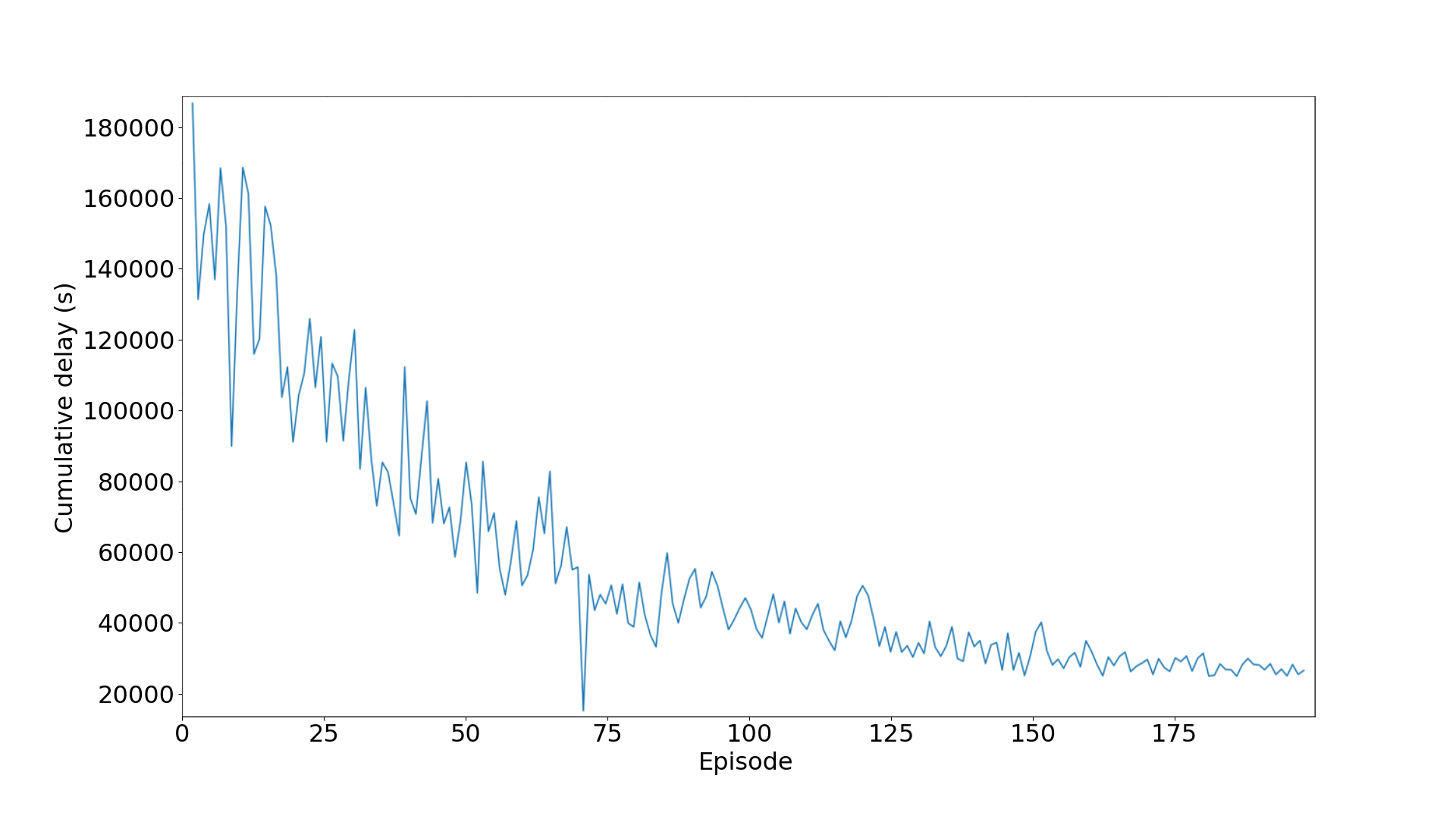}}
    \hfil
    \subfloat[Queue Lengths (average) at the intersection\label{fig:queue3}]{\includegraphics[height=14em,width=.45\linewidth]{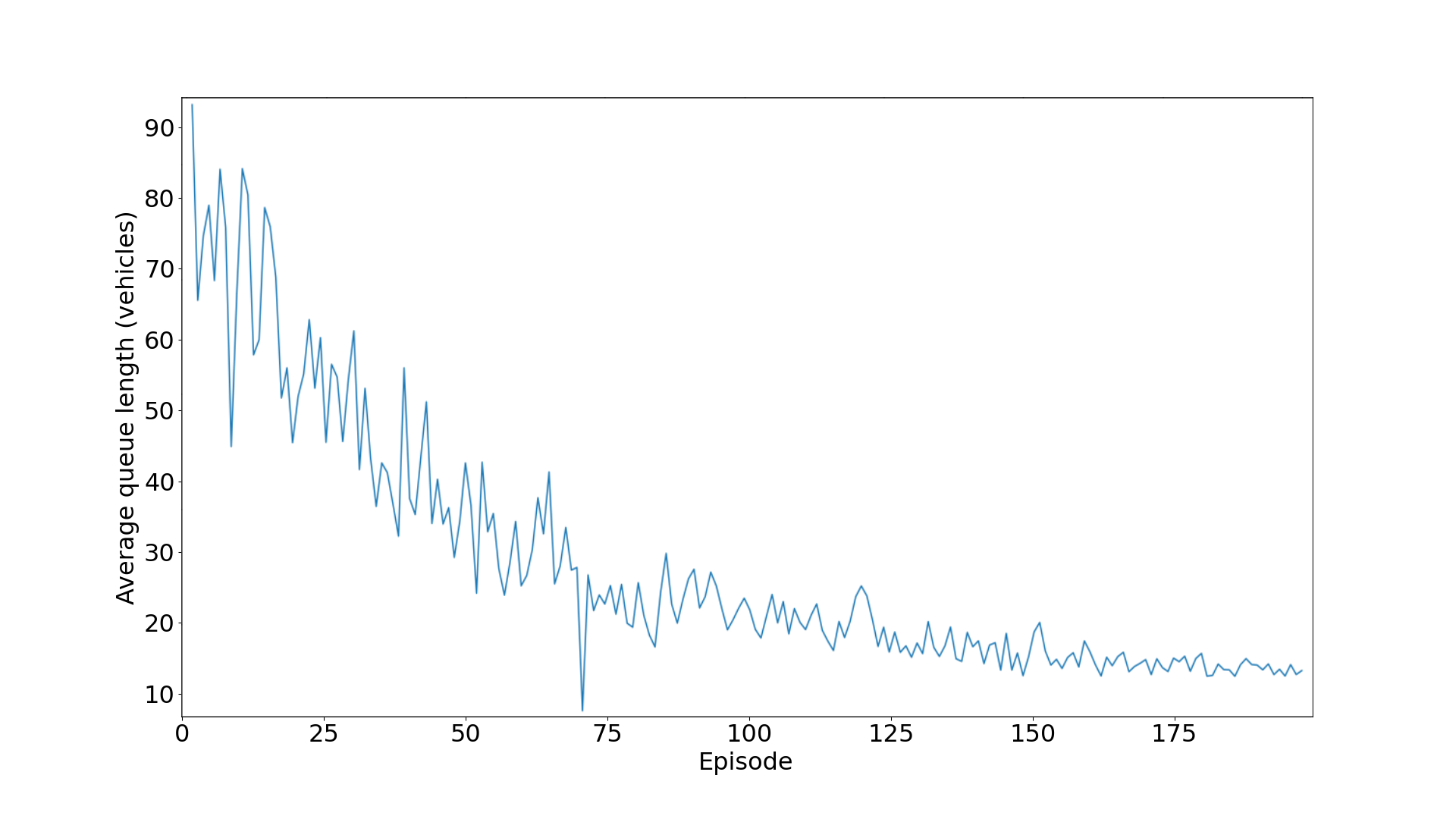}}

    \subfloat[Cumulative Negative Reward \label{fig:reward3}]{\includegraphics[height=14em,width=.45\linewidth]{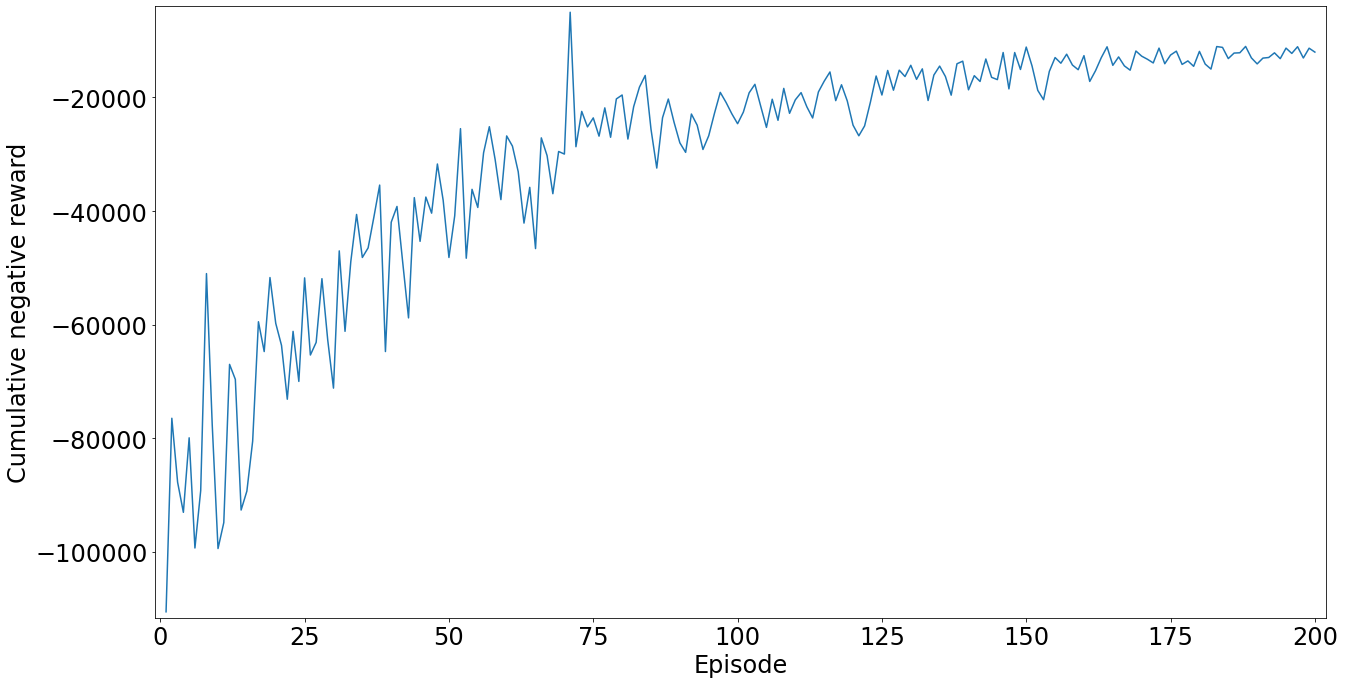}}
    \caption{Results of average delay, queue lengths and rewards for NN3}
    \label{fig:nn model3}
\end{figure*} 

\section{Results and Discussion}
 Our results are based on three phases. In first phase, we ran our simulation with the configuration using the conventional setting (pre-timed fixed) of traffic lights and recorded the outputs. The second phase shows the implementation of intelligent traffic lights using DRL for controlling traffic and its outputs. In the third phase, we used rerouting with intelligent traffic lights and recorded the results. Comparisons are shown at the end of this section. Now we will further explore the results of all these phases in detail.     

\subsection{Using pre-timed fixed signals:}
In this phase, we used pre-timed fixed signals for our experiments. Without implementing any intelligent traffic lights, we ran 4000 vehicles in the simulation and observed the traffic conditions. After few seconds, it was observed that vehicles at the intersection and behind the intersection got stuck due to the massive congestion on the intersection. The waiting time of the vehicles and delay increased with longer queue lengths. The system configuration and settings used for all experiments are the same. Results of experiments of 1st phase are shown in the 1st bar of Fig. \ref{fig:Performance graph}. The green bar shows that 4000 vehicles take almost 7392 (s) to reach the destination due to congestion on the roads and the intersection.   

\begin{figure*}[ht]
  \centering
    \subfloat[Comparisons for Cumulative Delay with different values of gamma\label{fig:combined delay}]{\includegraphics[height=16em,width=.45\linewidth]{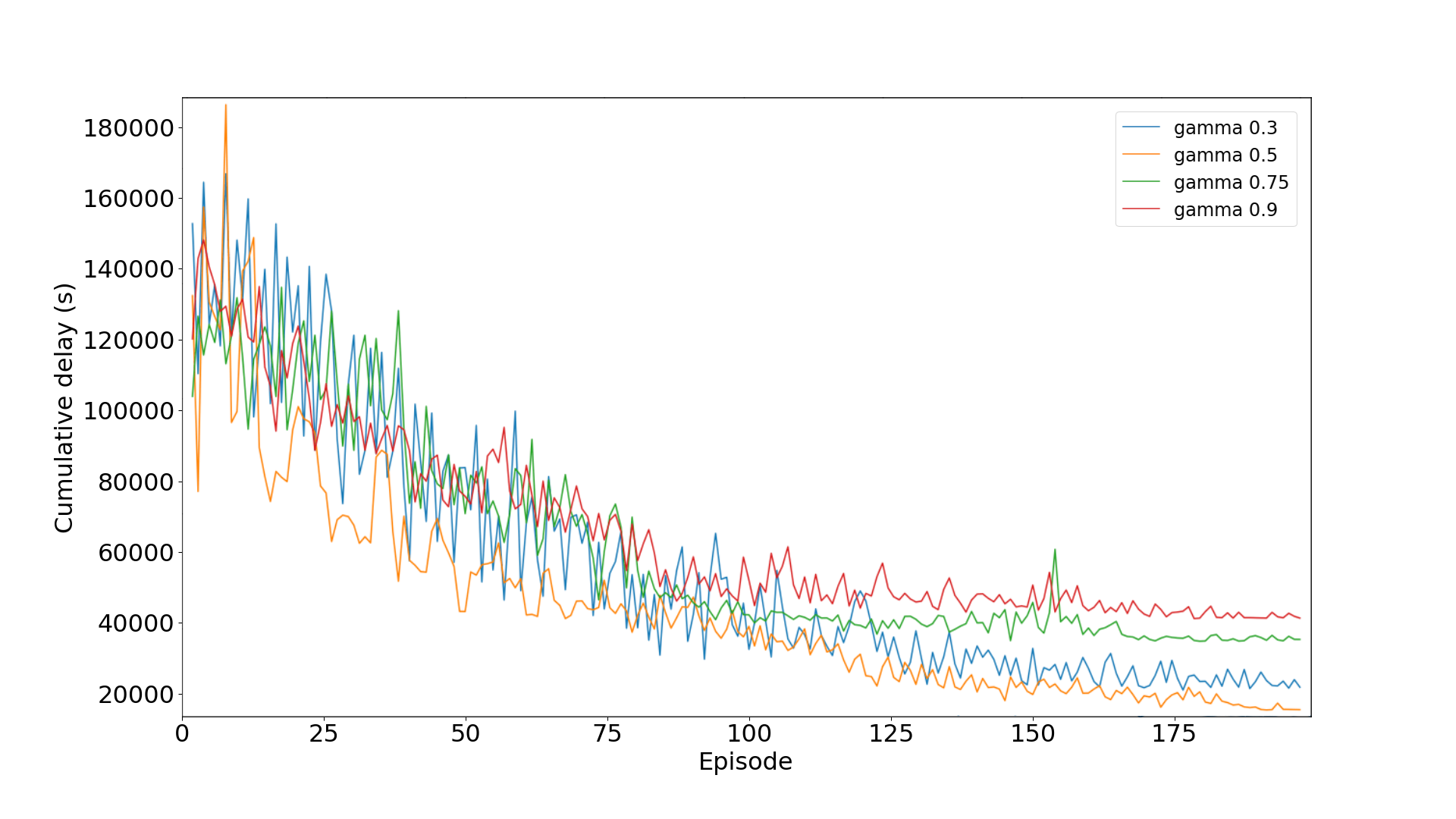}}
    \hfil
    \subfloat[Comparisons for Average Queue Length with different values of gamma\label{fig:combined queue}]{\includegraphics[height=14em,width=.45\linewidth]{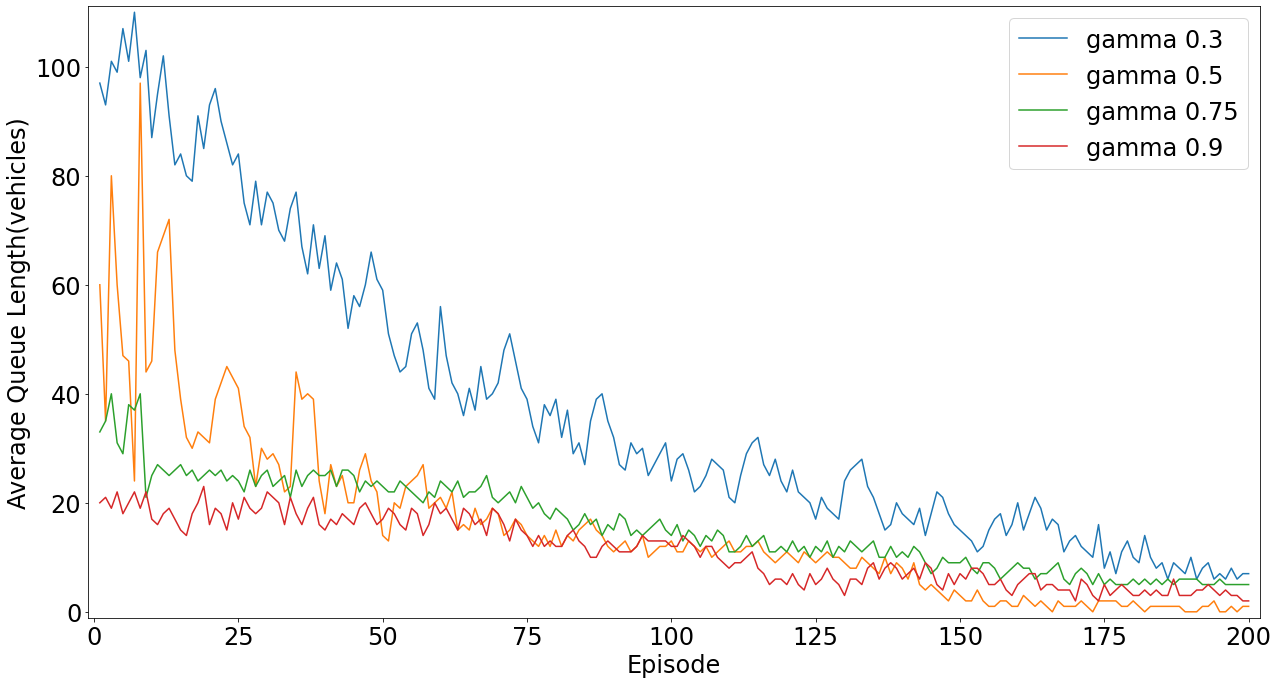}}

    \subfloat[Comparisons for Cumulative Negative Reward with different values of gamma\label{fig:combined reward}]{\includegraphics[height=14em,width=.45\linewidth]{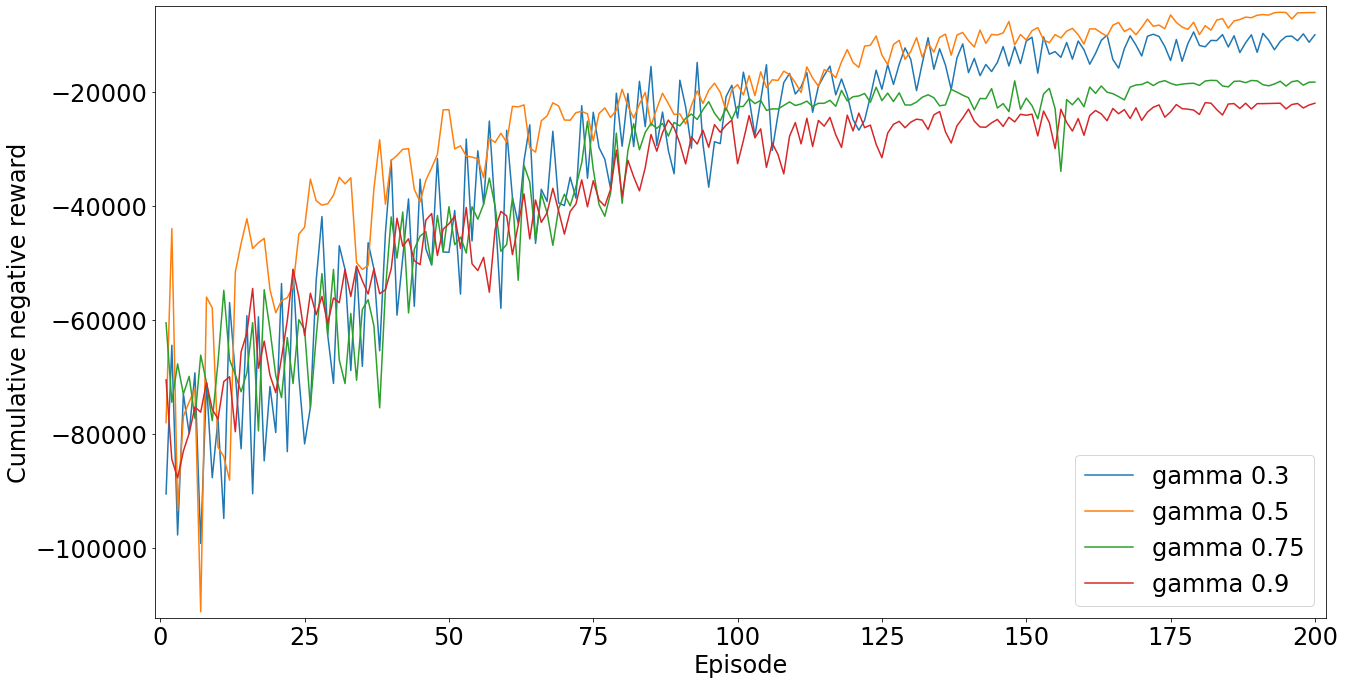}}
    \caption{Results of gamma value comparisons}
    \label{fig:gamma comparisons}
\end{figure*}

\subsection{Using RL-based controlled traffic lights}
In this phase, we used a policy-based DRL approach for traffic light control. Here, our agent consider the present condition of road's intersection from the data coming from detectors, sensors and data from other infrastructural elements. Then it computes the vehicle's wait time at intersections, queue lengths and takes appropriate action according to the situation. The signal that should be turned green depends on the computed wait time. Our agent measures the average wait time at the intersection and turns the signal on for that lane whose waiting time is more. The main purpose of our agent is to reduce the queue length, negative reward and cumulative delay of the vehicles. 
\par The Fig.\ref{fig:nn architecture} presents the architecture of Neural Network (NN) we used for our experiments. We used five-layered neural network models with different neurons on hidden layers, i.e. 200, 400 and 600 neurons. The input layer has 80 states, while the output layer has 4 states. The hidden layer of our neural network contains 200 neurons named  NN1, NN2 with 400 neurons, and NN3 with 600 neurons.
\par In all these three models, we can see a general trend that as the agent starts to learn about the environment with the passage of epochs, the commutative delay starts to decrease as well as the queue length. In contrast, the reward starts to increase, i.e. negative reward tends towards zero. In Fig.\ref{fig:nn model1}(a), Fig.\ref{fig:nn model2}(a), Fig.\ref{fig:nn model3}(a) we can observe that cumulative wait time (delay) starts decreasing as number of episodes increases. The traffic rate is the same on all lanes, and the average waiting time is measured at the end of each episode. 
 Similarly in Fig.\ref{fig:nn model1}(b), Fig.\ref{fig:nn model2}(b), Fig.\ref{fig:nn model3}(b) we can see that the average queue length of vehicles also starts decreasing which shows the congestion reduction on the intersection. As the agents learns the environment, it goes on managing traffic signals more properly and efficiently, because of which traffic flow is improved, and queues are reduced. In Fig.\ref{fig:nn model1}(c), Fig.\ref{fig:nn model2}(c), Fig.\ref{fig:nn model3}(c), we can see that total negative reward starts decreasing as agent learns more about the environment. After performing any action, the agent gets a reward (positive or negative), and our policy is updated based on that reward. In our case, we are using the negative reward. The maximum reward obtained shows that this is our optimal policy. In Fig.\ref{fig:nn model1}, Fig.\ref{fig:nn model2}, Fig.\ref{fig:nn model3}, we can see that after 175th episode there is no notable change in reward so we can say that after 175th episode we got our optimal policy.     
\par
\begin{figure*}[ht]
    \centering
    \includegraphics[width= 12cm,height=6cm]{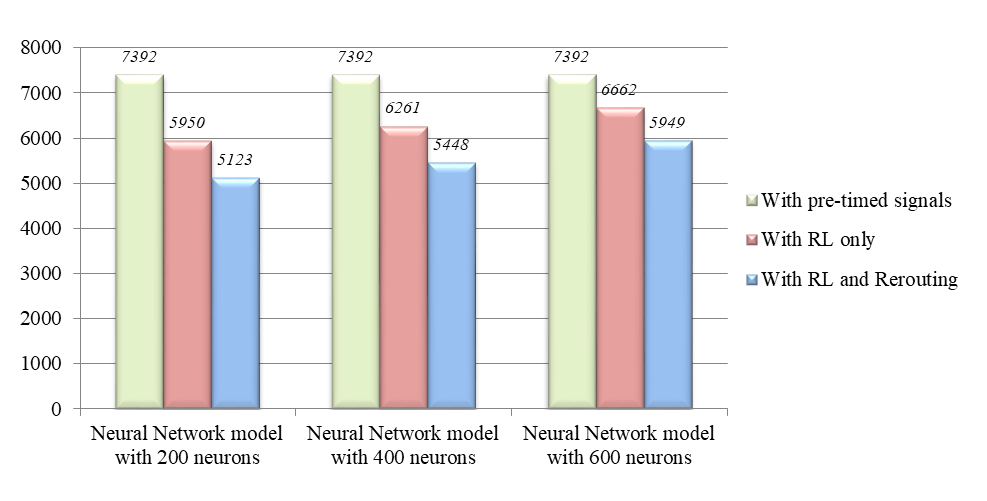}
    \caption{\textsf {Performance of Proposed Approach}}
    \label{fig:Performance graph}
\end{figure*}

 The red bar in Fig.\ref{fig:Performance graph} shows almost 20\% improvement in results with neural network model 1 when reinforcement learning-based traffic lights are applied on road networks. Similarly, there is 16\% and 10\% improvements in results while using neural network model 2 and 3 respectively. Here we can notice that when we increased neurons in hidden layers, the performance is degrading, so we can say the neural network model 1 is performing best for our case. There are no hard and fast rules for choosing the hidden layer's neurons; it depends on systematic experimentation.

\subsection{ Using Reinforcement based traffic lights with Rerouting}
In this phase of experiments, we used a combination of techniques for traffic congestion management. On traffic intersection, we used deep reinforcement learning-based traffic lights optimization. For the vehicles that are still not reached at intersection, and stuck in the congestion due to the long queue lengths at the Intersection, we rerouted them to alternate paths. This technique helped in two ways: first, the vehicles rerouted to alternate paths could get their destination without waiting for long time at the congested intersection and taking paths that are a bit lengthier but mostly less congested. Secondly, congestion was reduced on the intersection and road behind the intersection due to the rerouting of vehicles as now most vehicles were rerouted towards other paths, and vehicles were not continuously coming towards the intersection. In other words, traffic is load-balanced and divided into many paths that would improve the traffic flow and reduce congestion. In Fig.\ref{fig:Performance graph}, the blue bar shows the improvement of results by 14\% using 1st NN, 13\%  using 2nd NN and 11\% using 3rd NN. Thus the total reduction in time by using reinforcement learning and rerouting on NN1 is 34\%, NN2 is 29\% and NN3 is 21\%. The best results are 34\% using NN1. 

\subsection{Changing Gamma Value}
We want to make some comparisons changing the gamma values on our neural networks and see the effects at this point of experiments.  We did experiments on 4 values of gamma i.e. 0.3, 0.5, 0.7 and 0.9. Fig.\ref{fig:gamma comparisons} shows the effect of changing these values. From the figure, we can notice that the gamma value 0.5 is better than other values. 
\begin{figure*}[h!]
    \centering
    \includegraphics[width= 12cm,height=6cm]{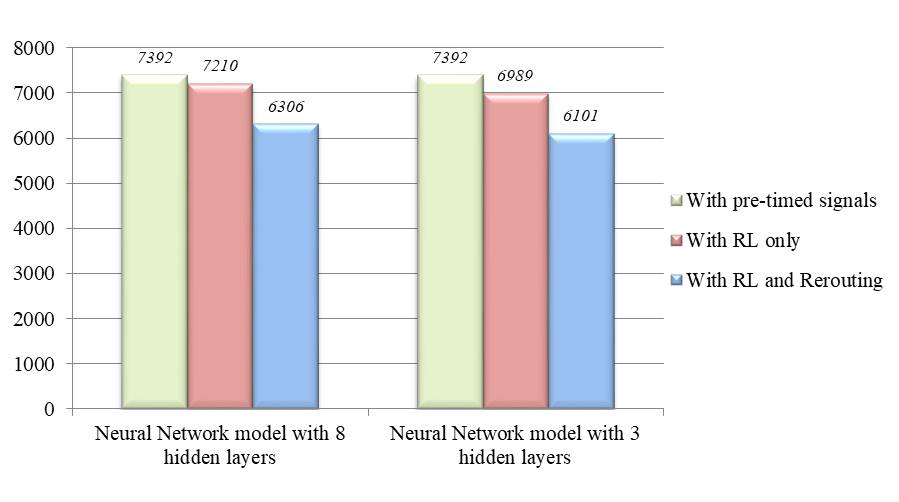}
    \caption{\textsf {Performance using deep and shallow models}}
    \label{fig:Performance graph2}
\end{figure*}

\subsection{Comparisons with Deeper vs Shallower Networks}
We made some comparisons in the last set of experiments using some deeper and shallower neural network models. A more deep neural network model has more hidden layers, while in a shallow neural network model, fewer hidden layers are used. In Fig.\ref{fig:deep} we used a deep NN-model with varying hidden layers between 5 to 8, and in Fig.\ref{fig:shallow} we used a shallow network with the number of hidden layers reduced from 5 to 3. The reason for these types of experiments is to show how a five-layered NN model is optimized. The red bar in the graph in Fig.\ref{fig:Performance graph2} shows that using a deep model with 8 hidden layers causes only a 3\% reduction in waiting time using RL  and 13\% reduction using rerouting with the same scenario. Similarly, a shallow network using only 3 hidden layers reduces the waiting time by 6\% when reinforcement learning is applied on traffic lights, and smart rerouting further reduces it to 13 \%. Thus the total simulation time is increased by 16\% as we increased the number of hidden layers compared to five layered neural network model that performs best for our system. Similarly, decreasing the number of hidden layers reduces the simulation time by 19\% and degrades our performance. These experiments suggest that for our approach, a five-layered neural network model performs the best. As traffic data is dynamic and non-linear, we have to find the best system configurations for our models so we perform different types of experimentation.

\section{Conclusion}
In this article, we proposed our approach for congestion management at road intersection and load balancing the vehicles in dynamically changing complex traffic environments. We applied the DRL approach to optimize the  intelligent traffic lights capable to take run-time decisions after analyzing the present state of the intersection. For this purpose, We used a deep neural network models with  different number of neurons and hidden layers. As a result, we figured out the optimum configuration for our neural network architecture, a five-layered NN-model with 200 neurons performs best for our system. We found that using DRL-based signal controllers improve traffic flow performance on intersections by 20\%. We then managed the traffic that is not yet reached at the intersection by rerouting it to alternate paths to avoid the congested intersections. It further improved our performance by 14\%. The total improvement in the performance is 34\%  which increased the efficiency, throughput and reduced the congestion, waiting time, delay and queue lengths of vehicles. The results shows the effectiveness of our proposed strategy to solve the traffic congestion problem while improving the flow management of traffic, especially when all vehicles are on their own without any driver present in them. The approach is equally helpful for regular vehicles and in mixed autonomy. The rerouting approach could be beneficial for emergency vehicles or special vehicles such as VIP vehicles or ambulances.
\newline At present, we are also working on extending this approach for more complex traffic scenarios using multi-intersection techniques and real-world road networks to help support the concept of futuristic intelligent cities.

\bibliographystyle{unsrt}
\bibliography{references.bib}

\end{document}